\documentclass{article} 
\usepackage{iclr2016_conference,times}
\usepackage{hyperref}
\usepackage{url}

\usepackage{booktabs,tabularx} 
\usepackage{graphicx}
\usepackage{mdwlist} 
\usepackage{multirow}

\usepackage{enumitem}

\usepackage[utf8]{inputenc}

\hypersetup{%
  pdftitle={Deep Feature Learning for EEG Recordings},
  pdfauthor={Sebastian Stober, Avital Sternin, Adrian M. Owen and Jessica A. Grahn},
  pdfkeywords={Convolutional Neural Networks, Deep Learning, Electroencephalography, Similarity Constraint Encoding},
  pdfstartview={FitH},
    colorlinks=true, linktocpage=true, pdfstartpage=1, pdfstartview=FitV,%
    breaklinks=true, pdfpagemode=UseNone, pageanchor=true, pdfpagemode=UseOutlines,%
    plainpages=false, bookmarksnumbered, bookmarksopen=true, bookmarksopenlevel=1,%
    hypertexnames=true, pdfhighlight=/O,
    urlcolor=blue, linkcolor=blue, citecolor=blue, 
    pdfcreator={pdfLaTeX},%
    pdfproducer={LaTeX with hyperref}%
}

\usepackage{acronym}
\acrodef{EEG}{electroencephalography}
\acrodef{EOG}{electrooculography}
\acrodef{MIIR}{music imagery information retrieval}
\acrodef{MIR}{music information retrieval}
\acrodef{BCI}{brain-computer interface}
\acrodef{STFT}{short-time Fourier transform}
\acrodef{SDA}{stacked denoising auto-encoder}
\acrodef{CAE}{convolutional auto-encoder}
\acrodef{MLP}{multilayer perceptron}
\acrodef{DBN}{deep belief net}
\acrodef{HMM}{hidden Markov model}
\acrodef{SGD}{stochastic gradient descent}
\acrodef{CNN}{convolutional neural network}
\acrodef{SSEP}{steady state evoked potential}
\acrodef{MEG}{magnoencephalography}
\acrodef{ERP}{event-related potential}
\acrodef{fMRI}{functional magnetic resonance imaging}
\acrodef{ANN}{artificial neural network}
\acrodef{MRR}{mean reciprocal rank}
\acrodef{SVM}{support vector machine}
\acrodef{PCA}{principle component analysis}
\acrodef{ICA}{independent component analysis}
\acrodef{MSRE}{mean square reconstruction error}
\acrodef{MCC}{mean channel correlation}

\usepackage{suffix}
\WithSuffix\newcommand\section*[1]{\noindent\textbf{#1:}} 

\usepackage{soul} 
\usepackage{todonotes} 
\newcommand{\addtodo}[1]{\addcontentsline{tdo}{todo}{\protect{#1}}#1}
\newcommand{\se}[1]{\emph{(\addtodo{\textcolor{blue}{Sebastian:~#1}})}}
\newcommand{\av}[1]{\emph{(\addtodo{\textcolor{orange}{Avital:~#1}})}}

\renewcommand{\hl}[1]{}
\renewcommand{\se}[1]{}
\renewcommand{\av}[1]{}

\newcommand{\tablesec}[1]{\multicolumn{9}{l}{#1}\\}

\title{Deep Feature Learning for EEG Recordings}

\author{
Sebastian Stober, Avital Sternin, Adrian M. Owen \& Jessica A. Grahn\\
The Brain and Mind Institute\\
University of Western Ontario\\
London, ON, Canada \\
\texttt{\{sstober,asternin,adrian.owen,jgrahn\}@uwo.ca} \\
}

\begin{document}

\maketitle

\begin{abstract}
We introduce and compare several strategies for learning discriminative features from electroencephalography (EEG) recordings using deep learning techniques.
EEG data are generally only available in small quantities,
they are high-dimensional with a poor signal-to-noise ratio,
and there is considerable variability between individual subjects and recording sessions.
Our proposed techniques specifically address these challenges for feature learning.
Cross-trial encoding forces auto-encoders to focus on features that are stable across trials. 
Similarity-constraint encoders learn features that allow to distinguish between classes by demanding that two trials from the same class are more similar to each other than to trials from other classes.
This tuple-based training approach is especially suitable for small datasets.
Hydra-nets allow for separate processing pathways adapting to subsets of a dataset and thus combine the advantages of individual feature learning (better adaptation of early, low-level processing) with group model training (better generalization of higher-level processing in deeper layers).
This way, models can, for instance, adapt to each subject individually to compensate for differences in spatial patterns due to anatomical differences or variance in electrode positions.
The different techniques are evaluated using the publicly available OpenMIIR dataset
of EEG recordings taken while participants listened to and imagined music.
\end{abstract}

\section{Introduction}

Over the last decade, deep learning techniques have become very popular in various application domains such as
computer vision, automatic speech recognition, natural language processing, 
and bioinformatics where they 
produce state-of-the-art results on various tasks. 
At the same time, there has been very little progress investigating the application of deep learning in cognitive neuroscience research, where these techniques could be used to analyze signals recorded with
\acf{EEG} -- a non-invasive brain imaging technique that relies on electrodes placed on the scalp to measure the electrical activity of the brain.
\ac{EEG} is especially popular for the development of \acp{BCI}, which work by identifying different brain states from the \ac{EEG} signal.

Working with \ac{EEG} data poses several challenges.
Brain waves recorded in the \ac{EEG} have a very low signal-to-noise ratio and the noise can come from a variety of sources. 
For instance, the sensitive recording equipment can easily pick up electrical line noise from the surroundings.
Other unwanted electrical noise can come from muscle activity, eye movements, or blinks.
Usually, only certain brain activity is of interest, and this signal needs to be separated from background processes.
\ac{EEG} lacks spatial resolution on the scalp with additional spatial smearing caused by the skull but it has a good (millisecond) time resolution to record both, slowly and rapidly changing dynamics of brain activity.
Hence, in order to identify the relevant portion of the signal, sophisticated analysis techniques are required that should also take into account temporal information.

This is where deep learning techniques could help.
For these techniques, training usually involves the usage of large corpora such as ImageNet 
or TIMIT. 
As \ac{EEG} data are high-dimensional\footnote{%
A single trial comprising ten seconds of EEG with 64 channels sampled at 100 Hz has already 64000 dimensions and the number of channels and the sampling rate of EEG recordings can be much higher than this.}
and complex, this also calls for large datasets to train deep networks for \ac{EEG} analysis and classification.
Unfortunately, there is no such abundance of \ac{EEG} data.
Unlike photos or texts extracted from the internet, \ac{EEG} data are costly to collect and generally unavailable in the public domain.
It requires special equipment and a substantial effort to obtain high quality data.
Consequently, \ac{EEG} datasets are only rarely shared beyond the boundaries of individual labs and institutes.
This makes it hard for deep learning researchers to develop more sophisticated analysis techniques tailored to this kind of data.

With this paper, we want to address several of these challenges.
We briefly review related work in \autoref{sec:related_work} and introduce our dataset in \autoref{sec:dataset}.
\autoref{sec:experiments} describes our proposed techniques and the experiments conducted to test them.
We discuss the results and draw conclusions in \autoref{sec:conclusions}.
Supplementary material in the appendix comprises further illustrations of learned features as well as details on the implementation of the proposed methods.

\section{Related Work}\label{sec:related_work}

A major purpose of the work presented in this paper is to help advance the state of the art of signal analysis techniques in the field of cognitive neuroscience.
In this application domain, the potential of deep learning techniques for neuroimaging has been demonstrated very recently by \cite{Plis2013Deep-learning-for-neuro} for functional and structural magnetic resonance imaging (MRI) data.
However, applications of deep learning techniques within cognitive neuroscience and specifically for processing \ac{EEG} recordings have been very limited so far.

\cite{
mirowski_classification_2009} 
applied \acp{CNN} for epileptic seizure prediction in EEG and intercranial EEG.
\cite{Wulsin2011Modeling-electroence} used \acp{DBN} to detect anomalies related to epilepsy in \ac{EEG} recordings 
by classifying individual ``channel-seconds'', i.e., one-second chunks from a single \ac{EEG} channel without further information from other channels or about prior values.
Their classifier was first pre-trained layer by layer as an auto-encoder on unlabelled
data, followed by a supervised fine-tuning with backpropagation on a much smaller labeled dataset.
They found that working on raw, unprocessed data (sampled at 256Hz) led to a classification accuracy comparable to hand-chosen features.
\cite{Langkvist2012Sleep-Stage-Classifi} similarly employed \acp{DBN} combined with \acp{HMM} to classify different sleep stages. 
Their data for 25 subjects comprised \ac{EEG} as well as recordings of eye movements and skeletal muscle activity.
Again, the data was segmented into one-second chunks.
Here, a \ac{DBN} on raw data showed a classification accuracy close to one using 28 selected features.

Furthermore, there have been some applications of \acp{CNN} for \acp{BCI}.
\cite{cecotti_convolutional_2008} used a special \ac{CNN} for classification of steady-state visual evoked potentials (SSVEPs) -- i.e., brain oscillation induced by visual stimuli.
The network integrated the Fourier transform between convolutional layers, which transformed the data from the time domain to a time-frequency representation.
Another \ac{CNN} for detecting P300 waves (a well established waveform in \ac{EEG} research) was described in \cite{cecotti_convolutional_2011}.
There has also been early work on emotion recognition from EEG using deep neural networks such as described by \cite{jirayucharoensak_eeg-based_2014} and \cite{zheng_eeg-based_2014}.
In our early work, we used \acp{SDA} and \acp{CNN} to classify EEG recordings of rhythm perception and identify their 
ethnic origin -- East African or Western -- (\cite{stober2014ismir})
as well as to distinguish individual rhythms (\cite{stober2014nips}).

\section{Dataset and Pre-Processing}\label{sec:dataset}

The OpenMIIR dataset (\cite{stober2015ismir}) is a public domain dataset of \ac{EEG} recordings taken during music perception and imagination.\footnote{%
The dataset is available at \url{https://github.com/sstober/openmiir}}
We collected this data from 10 subjects who listened to and imagined 12 short music fragments -- each 7s--16s long -- taken from well-known pieces.
These stimuli were selected from different genres and systematically span several musical dimensions such as meter, tempo and the presence of lyrics as shown in \autoref{tab:stimuli_information} in the appendix.
This way, various retrieval and classification scenarios can be addressed.

All stimuli 
were normalized in volume and kept as similar in length as possible with care taken to ensure that they all contained complete musical phrases starting from the beginning of the piece. 
The pairs of recordings for the same song with and without lyrics were tempo-matched.
The stimuli were presented to the participants in several conditions while we recorded EEG. 
For the experiments described in this paper, we only focus on the perception condition, where participants were asked to just listen to the stimuli.
The presentation was divided into 5 blocks that each comprised all 12 stimuli in randomized order.
In total, 60 perception trials 
were recorded per subject.

EEG was recorded with a BioSemi Active-Two system using 64+2 EEG channels at 512\,Hz. 
Horizontal and vertical \ac{EOG} channels were used to record eye movements. 
The following common-practice pre-processing steps were applied to the raw EEG and EOG data using the MNE-python toolbox by \cite{gramfort_meg_2013} to remove unwanted artifacts.
We removed and interpolated bad EEG channels (between 0 and 3 per subject) identified by manual visual inspection.\footnote{%
The removed and interpolated bad channels are marked in the topographic visualizations shown in \autoref{fig:topoplots}.
}
The data was then filtered with a bandpass keeping a frequency range between 0.5 and 30\,Hz.
This also removed any slow signal drift in the EEG.
To remove artifacts caused by eye blinks, we computed independent components using extended Infomax \ac{ICA} as described by \cite{lee_independent_1999} and semi-automatically removed components that had a high correlation with the EOG channels.
Afterwards, the 64 EEG channels were reconstructed from the remaining independent components without reducing dimensionality.
Furthermore, the data of one participant was excluded at this stage because of a considerable number of trials with movement artifacts due to coughing.
Finally, all trial channels were additionally normalized to zero mean and range $[-1,1]$.

\section{Experiments}\label{sec:experiments}

Using the EEG dataset described in the previous section, we would like to learn discriminative features that can be used by a classifier to distinguish between the different music stimuli.
Ideally, these feature should also allow interpretation by cognitive neuroscientists to facilitate findings about the underlying cognitive processes.
In our previous experiments with EEG recordings of rhythm perception, \acp{CNN} showed promising classification performance but the learned features were not easy to interpret (\cite{stober2014nips}).

For the experiments described here, the following general implementation conventions applied:
All convolutional layers used the sigmoid tanh nonlinearity because its output naturally matches the value range of the network inputs ([-1,1]) and thus facilitates easier interpretation of the activation values.
Furthermore, bias terms were not used.
Convolution was always solely applied along the time (samples) axis.\footnote{%
Beyond the scope of this paper, convolution could be applied in the spatial or frequency domain.
}
For the classifiers, we used a DLSVM output layer employing the hinge loss as described by \cite{tang_deep_2013} with an implementation based on the one provided by 
Kastner.\footnote{%
\url{https://github.com/kastnerkyle/pylearn2/blob/svm_layer/} 
}
This generally resulted in a better classification performance than the commonly used Softmax in all our previous experiments.
For the \acp{CAE}, our implementation of the de-convolutional layers has been derived from the code for generative adversarial nets by \cite{goodfellow_generative_2014}.\footnote{\url{https://github.com/goodfeli/adversarial}}
Stochastic gradient descent with batches of 128 trials was used for training.
During supervised training, we applied Dropout regularization (\cite{hinton2012dropout}) and a learning rate momentum. 
During unsupervised pre-training, we did not use Dropout as the expected benefit is much lower here and does not justify the increase in processing time.
Generally, the learning rate was set to decay by a constant factor per epoch.
Furthermore, we used a L1 weight regularization penalty term in the cost function to encourage feature sparsity.

To measure classification accuracy, we used the trials of the third block of each subject as test set.
This set comprised 108 trials (9 subjects x 12 stimuli x 1 trial).
The remaining 432 trials (9 subjects x 12 stimuli x 4 trials) were used for training and model selection.
For supervised training, we employed a 9-fold cross-validation scheme by training on the data from 8 subjects (384 trials) and validating on the remain one (48 trials).
This approach allowed us to additionally estimate the cross-subject performance of a trained model.\footnote{%
There is still a model selection bias because the cross-subject validation set is used for early stopping.
}
For hyper-parameter selection, we employed the Bayesian optimization technique described by \cite{Snoek2012Practical-Bayesian-O} which has been implemented in the \emph{Spearmint} library.\footnote{%
\url{https://github.com/JasperSnoek/spearmint}
}
We limited the number of jobs for finding optimal hyper-parameters to 100 except for the baseline at 64\,Hz for which we ran 300 jobs.
Each job comprised training the 9 fold models for 50 epochs and selecting the model with the lowest validation error for each fold.
We compared two strategies for aggregating the separate fold models -- averaging the model parameters over all fold models (``avg'') or using a majority vote (``maj'').
The accuracy on the test set is reported for these aggregated models.

In the experiments described in the following, we focused on pre-training the first \ac{CNN} layer.
The proposed techniques can nevertheless also be applied to train deeper layers and network structures different from \acp{CNN}.
Once pre-trained, the first layer was not changed during supervised classifier training to obtain a measurement of the feature utility for the classification task.
Furthermore, this also resulted in a significant speed-up of training -- especially at the high sampling rate of 512\,Hz.
Except for the pre-training method described in \autoref{sec:learning_basic}, which uses full-length trials, all trials were cut off at 6.9\,s, the length of the shortest stimulus, which resulted in an equal input length. 

For comparison, we additionally trained a linear support vector machine classifier (SVC) on top of the learned features using Liblinear (\cite{fan2008liblinear}).
We also tested polynomial kernels, but this did not lead to an increase in classification accuracy.
For the SVC, the optimal value for the parameter C that controls the trade-off between the model complexity and the proportion of non-separable training instance was determined through a grid search during cross-validation.

\subsection{Supervised CNN Training Baseline}

In order to establish a baseline for the OpenMIIR dataset, we first applied plain supervised classifier training.
We considered \acp{CNN} with two convolutional layers using raw EEG as input, which was either down-sampled to 64 Hz or kept at the original sampling rate of 512 Hz, as well as respective SVCs for comparison.
The higher sampling rate offers better timing precision at the expense of increasing the processing time and the memory requirement.
We wanted to know whether using data at the full rate of 512 Hz could by justified by increasing our classification accuracy.
We conducted a search on the hyper-parameter grid optimizing solely structural network parameters and the learning rate.
Results are shown in \autoref{tab:results} (B and E).
The baseline results give us a starting point against which we can compare the results obtained through our proposed feature learning approaches.

\subsection{Learning Basic Common Signal Components}\label{sec:learning_basic}

No matter how much effort one puts into controlling the experimental conditions during \ac{EEG} recordings,
there will always be some individual differences between subjects and between recording sessions.
This can make it hard to combine recordings from different subjects to identify general patterns in the \ac{EEG} signals.
A common way to address this issue is to average over many very short trials such that differences cancel out each other.
When this is not feasible because of the trial length or a limited number of trials, 
an alternative strategy is to derive signal components from the raw \ac{EEG} data hoping that these will be stable and representative across subjects.
Here, \ac{PCA} and \ac{ICA} 
are well-established techniques.
Both learn linear spatial filters, whose channel weights can be visualized by topographic maps as shown in \autoref{fig:topoplots}.
The first two columns contain the signal components learned with PCA and extended Infomax ICA (\cite{lee_independent_1999}) on the concatenated perception trials
whereas the remaining components have been obtained using a \acfp{CAE} with 4 filters.

Using \acp{CAE}, allows us to learn individually adapted components that are linked between subjects.
To this end, a \ac{CAE} is first trained on the combined recordings from the different subjects to identify the most common components.
The learned filter weights are then used as initial values for individually trained auto-encoders for each subject.
This leads to adaptations that reflect individual differences.
Ideally, these are small enough such that the relation to the initial common components still remains as can be seen in \autoref{fig:topoplots}.
Optionally, a regularization term can be added to the cost function to penalize weight changes, but this was not necessary here. 

We measured reconstruction error values similar to those obtained with PCA and substantially lower than for ICA.
This demonstrates the suitability for general dimensionality reduction with the additional benefit of obtaining a common data representation across subjects that can accommodate individual differences.
Using the filter activation of either the global or the individually adapted \acp{CAE} shown in \autoref{fig:topoplots} as common feature representation, we constructed SVC and \ac{CNN} classifiers.
The latter consisted of another convolutional layer and a fully connected output layer with hinge loss.
The obtained accuracies after hyper-parameter optimization are shown in \autoref{tab:results} (IDs G and I).

\subsection{Cross-Trial Encoding}\label{sec:CTE}

Aiming to find signal components that are stable across trials of the same class, we changed the training scheme of the \ac{CAE} to what we call \emph{cross-trial encoding}.
Instead of trying to simply reconstruct the input trial, the \ac{CAE} now had to reconstruct a different trial belonging to the same class.\footnote{%
Under a strict perspective, trying to reconstruct a different trial may no longer be considered as \emph{auto}-encoding. 
However, we rather see the term as a reference to the network architecture, which is still the same as for regular auto-encoders.
Only the training data has changed.
}
This strategy can be considered as a special case of the generalized framework for auto-encoders proposed by \cite{wang_generalized_2014}.\footnote{%
They similarly define a ``reconstruction set'' (all instances belonging to the same class) but use a different (and much simpler) loss function based on linear reconstruction and do not consider convolution.
}
Given $n_C$ trials for a class $C$, $n_C^2$ or $n_C (n_C - 1)$ pairs of input and target trials can be formed depending on whether pairs with identical trials are included.
This increases the number of training examples by a factor depending on the partitioning of the dataset with respect to the different classes.
As for the basic auto-encoding scheme, the training objective is to minimize a reconstruction error.
In this sense, it is unsupervised training but the trials are paired for training using knowledge about their class labels.
We found that using the distance based on the dot product worked best as reconstruction error.

\begin{figure}[t] 
  \begin{center}
	\includegraphics[width=\textwidth,keepaspectratio=true]{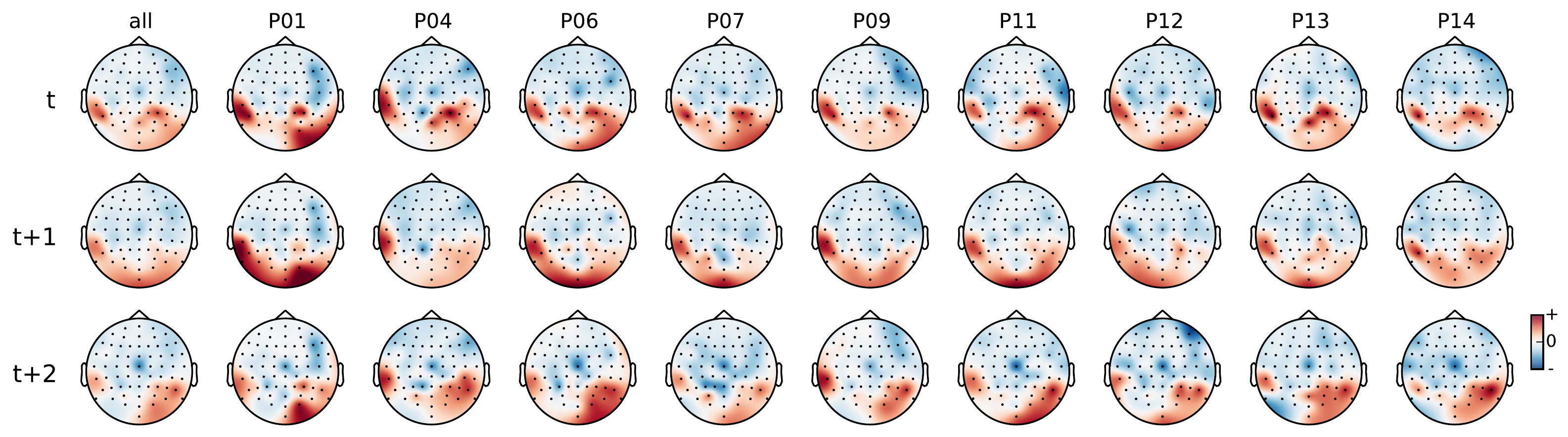}
   \\\vspace{-0.8em}
    \caption{%
A spatio-temporal filter over all 64 EEG channels and 3 samples after pre-training using cross-trial encoding.
Left column: Global filter after training on within-subject trial pairs (stage 1). 
Remaining columns: Individual filters (hydra-net) after training with cross-subject trials (stage 2).
The output of this filter was used in the classifiers with ID M listed in \autoref{tab:results}.
}
    \label{fig:topoplots_cte_oc1kw3}
  \end{center}
  \vspace{-1em}
\end{figure}

We split the training process into two stages.
In stage 1, trials were paired within subjects whereas in stage 2, pairs across subjects were considered as well.
To allow the \ac{CAE} to adapt to individual differences between the subjects in stage 2, we introduced a modified network structure called \emph{hydra-net}.
Hydra-nets allow to have separate processing pathways for subsets of a dataset.
This makes it possible to have different weights -- depending on trial meta-data -- in selected layers of a deep network. 
Such a network can, for instance, adapt to each subject individually. 
With individual input layers, the structure can be considered as a network with multiple ``heads'' -- hence the reference to Hydra.
Here, we applied this approach at both ends of the \ac{CAE}, i.e., the encoder filters were selected based on the input trial's subject whereas the decoder filters were chosen to match the target trial's subject.
Encoder and decoder weights were tied within subjects.
\autoref{fig:topoplots_cte_oc1kw3} shows the result learned by a \ac{CAE} with a single filter of width 3 after stage 1 and 2.
Further examples are shown in \autoref{sec:CTE_more}.
Classification results were obtained in the same way as in \autoref{sec:learning_basic} and are shown in \autoref{tab:results} (K--O).

\subsection{Similarity-Constraint Encoding}\label{sec:sim-constraint_encoding}

Cross-trial encoding as described above aims to identify features that are stable across trials and subjects.
Ideally, however, features should also allow us to distinguish between classes
but this is not captured by the reconstruction error used so far.
In order to identify such features, we propose a pre-training strategy called \emph{similarity-constraint encoding}
in which the network is trained to encode relative similarity constraints.
As introduced by \cite{schultz_learning_2004}, a relative similarity constraint $(a,b,c)$ describes a relative comparison of the trials $a$, $b$ and $c$ in the form ``$a$ is more similar to $b$ than $a$ is to $c$.''
Here, $a$ is the reference trial for the comparison.
There exists a vast literature on using such constraints to learn similarity measures in various application domains.
Here, we use them to define an alternative cost function for learning a feature encoding.
To this end, we combine all pairs of trials $(a,b)$ from the same class (as described in \autoref{sec:CTE}) with all trials $c$ from other classes demanding that $a$ and $b$ are more similar.

All trials within a triplet that constitutes a similarity constraint are processed using the same encoder pipeline.
This results in three internal feature representations.
Based on these, the reference trial is compared with the paired trial and the trial from the other class resulting in two similarity scores.
Here, we used the dot product as similarity measure because this matched the computation performed later during supervised training in the classifier output layer.
The output layer of the similarity constraint encoder finally predicts the trial with the highest similarity score without further applying any additional transformations.
The whole network can be trained like a common binary classifier, minimizing the error of predicting the wrong trial as belonging to the same class as the reference.
Technical details and a schematic of our similarity-constraint encoding approach can be found in \autoref{sec:SCE_details}.

This strategy is different from the one proposed by \cite{yang_learning_2014} 
to jointly learn features together with similarity metrics.
In particular, they used pairs for training and predicted whether these were from the same or different classes.
Their approach also required balancing two cost functions (reconstruction vs. discrimination). 
We hypothesize that our relative comparison approach leads to smoother learning.
Each fulfilled constraint only makes a small local contribution to the global structure of the similarity space.
This way, the network can more gradually adapt compared to a scenario where it would have to directly recognize the different classes based on a few training trials. 
Optionally, the triplets can be extended to tuples of higher order by adding more trials from other classes.
This results in a gradually harder learning task because there are now more other trials to compare with.
At the same time, each single training example comprises multiple similarity constraints, which might speed up learning.
In the context of this paper, we focus only on triplets.

\begin{figure}
  \begin{center}
     \includegraphics[width=0.8\textwidth,keepaspectratio=true]{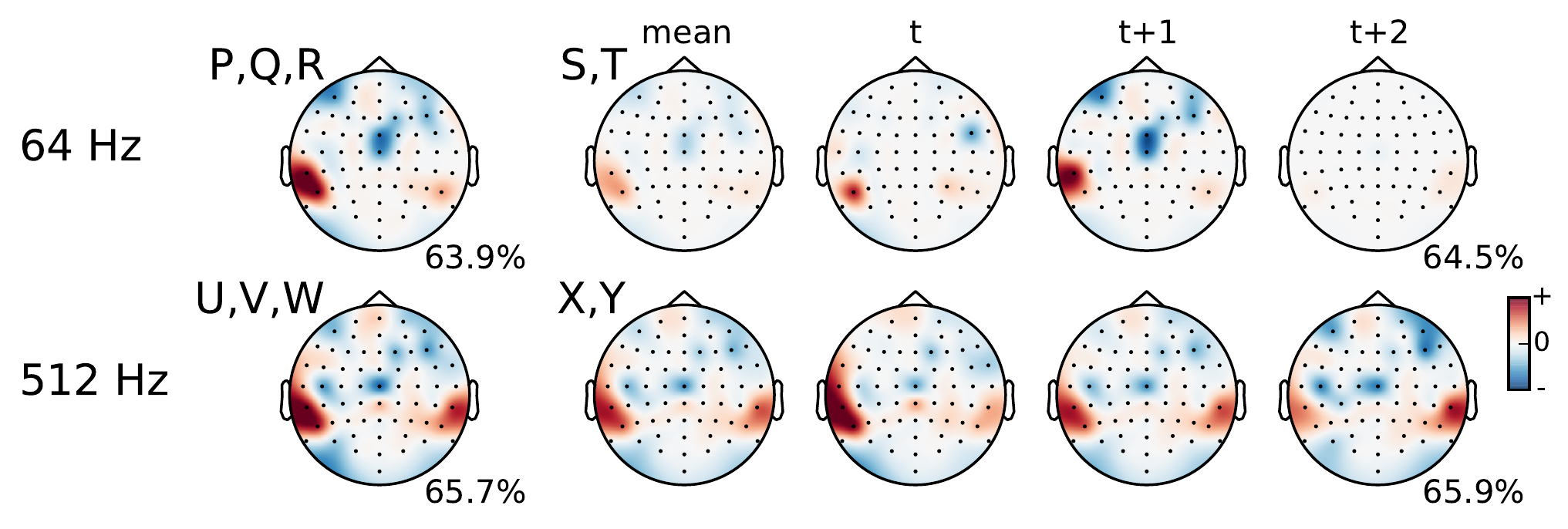}
   \\\vspace{-0.8em}
    \caption{%
Filters learned by similarity-constraint encoding for EEG sampled at 64\,Hz (top row) and 512\,Hz (bottom row).
Left: Simple spatial filters with width 1.
Right: Spatio-temporal filters with width 3. 
The mean over all time steps is shown additionally for comparison.
Letters P--Y refer to the classifier IDs in \autoref{tab:results} and \autoref{tab:result_details} that use the respective filter as their first layer.
The percentage of correctly classified training triples is shown at the lower right of each filter.
}
    \label{fig:topoplots_sce}
  \end{center}
  \vspace{-1em}
\end{figure}

\autoref{fig:topoplots_sce} shows filters learned by similarity-constraint encoding for different structural configurations.
They look very different from the results obtained earlier.
Remarkably, the same pattern emerged in all tested configurations.
Using more than one filter generally did not lead to improvements in the constraint encoding performance.
A kernel width of more than 3 samples at 64\,Hz sampling rate did not lead to different patterns as channel weights for further time steps remained close to zero.
The classification accuracies of using these pre-trained features in SVC and \ac{CNN} classifiers with optimized hyper-parameters are reported in \autoref{tab:results} (IDs Q--Y).

\begin{table}[t]
\setlength{\tabcolsep}{3.5pt}  
\renewcommand{\arraystretch}{1.1}
\centering
\caption{Classification accuracy measured on the test set using the different feature learning techniques described earlier
in combination with SVC and CNN classifiers. 
Convolution parameters for feature computation are shown as the number of filters times the filter width in samples.
Classifier hyper-parameters were optimized during cross-validation.
The ID refers to the respective row in \autoref{tab:result_details}, which provides more details on the CNN parameters for the different configurations.
 }
\label{tab:results}
{\footnotesize  
\begin{tabular}{l c   c c c  c }
\toprule
Feature Learning Technique													& Filters	& SVC		& CNN (avg)	& CNN (maj)		& ID	\\
\midrule
None (baseline using raw EEG at 64\,Hz)								&			& 14.8\%	& 18.5\% 	& 26.9\%			& B 	\\
None (baseline using raw EEG at 512\,Hz)								&			& 14.8\%	& 18.5\% 	& 19.4\%			& E 	\\
Common Components (64\,Hz, global filters)							& 4x1		& 17.6\%	& 17.6\% 	& 18.5\%			& G 	\\
Common Components (64\,Hz, individually adapted filters)		& 4x1		& 19.4\%	& 20.4\% 	& 20.4\%			& I 	\\
Cross-Trial Encoder (64\,Hz, individually adapted filters)			& 1x1 	& 16.7\%	& 20.4\% 	& 19.4\%			& K	\\
																					& 1x3 	& 23.1\%	& 23.1\% 	& 22.2\%			& M	\\
																					& 4x1 	& 20.4\%	& 22.2\% 	& 22.2\%			& O	\\
Similarity-Constraint Encoder (64\,Hz)									& 1x1 	& 20.4\%	& 25.0\% 	& 27.8\%			& Q	\\
																					& 1x3 	& 24.1\%	& 20.4\% 	& 18.5\%			& T	\\
Similarity-Constraint Encoder (512\,Hz)									& 1x1 	& 22.2\%	& 22.2\% 	& 23.1\%			& V	\\
																					& 1x3 	& 23.1\%	& 27.8\% 	& 26.9\%			& Y	\\
\bottomrule
\end{tabular}
}
\vspace{-1em}
\end{table}

\section{Discussion}\label{sec:conclusions}

Trying to determine which music piece somebody listened to based on the \ac{EEG} is a challenging problem.
Attempting to do this with a small training set, makes the task even harder.
Specifically, in the experiments described above, we trained classifiers
for a 12-class problem (recognizing the 12 music stimuli listed in \autoref{tab:stimuli_information})
on the combined data from 9 subjects
with an input dimensionality of 28,160 (at 64\,Hz) or 225,280 (at 512\,Hz) respectively
given only 4 training examples per class and subject.

Remarkably, all \ac{CNN} classifiers listed in \autoref{tab:results} had an accuracy that was significantly above chance.
Even for model G, the value of 17.59\% was significant at p=0.001.
Significance values were determined by using the cumulative binomial distribution to estimate the likelihood of observing a given classification rate by chance.
It is likely that the classification accuracy will slightly increase when the pre-trained filters of the first layer are allowed 
to change during supervised training of the full \ac{CNN}.
We did not consider this here because we wanted to analyze the impact of the pre-trained features.
There appears to be a ceiling effect as the median cross-validation accuracy did not exceed 40\%.
Examining the cross-validation performance for the individual subjects, however,
indicates that cross-subject accuracies above 50\% are possible for some of the subjects when using individual models.
We plan to investigate this possibility in the future.
For this paper, we focused on training with the combined data from all subjects using new techniques to compensate for individual differences
as we expected this to result in more robust features.

For CNN model aggregation, averaging the filters of the 9 cross-validation fold models (``avg'') worked surprisingly well in general and did not result in a significantly different performance compared to using a majority vote (``maj'').
A single average CNN also has the advantage that it is much easier to analyze than the 9 individual voting models.

Comparing the baseline classifiers (using raw EEG) with those relying on the learned features as listed in \autoref{tab:results} using a z-test, we found a significant improvement at p=0.05 for the SVC model T.
For the average CNN model Y, the improvement over the respective baseline was significant at p=0.061.
Both classifiers relied on features learned by similarity-constraint encoding.
For the SVC models M and Y, we obtained a p-value of 0.068.
Due to the rather small test set size (n=108), many performance differences were only significant at higher p-values.
There appears to be a small performance advantage in using the full sampling rate of 512\,Hz as indicated by the scatter plots of the model performance during hyper-parameter optimization shown in \autoref{sec:scatterplots}.
Using similarity-constraint encoding to pre-train a simple spatial filter for the first \ac{CNN} layer that only aggregates the 64 EEG channels into a single waveform
not only significantly improved the classification accuracy but also reduced the time needed for hyper-parameter optimization from over a week for the full \ac{CNN} to a few hours.
This improvement makes working at the full sampling rate feasible.

In our experiments, we used similarity-constraint encoding to learn global filters.
These filters could be further adapted to the individual subjects through our hydra-net approach.
We believe that this combination has a high potential to further improve the pre-trained filters and the resulting classification accuracy.
However, technical limitations of our current implementation cause a significant increase in processing time for this setting (details can be found in \autoref{sec:implementation}).
We are currently working on a different implementation to resolve this issue.

Amongst the best \ac{CNN} classifiers, which are listed in \autoref{tab:result_details} with more details, models R and W stand out for their simplicity and accuracy that is on par with much bigger models.
Their improvement over the baseline is significant at p=0.1 and p=0.05 respectively.
Remarkably, model R does effectively not even use the second convolutional layer.
We interpret this as an indicator for the quality of the pre-trained features.
Both models are simple enough to allow for interpretation of the learned features by domain experts.
Visualizations can be found in \autoref{sec:model_visualizations}.
The visualizations of the layer 1 filters indicate which recording electrodes are most important to the classifier.
The electrodes within the dark red areas that appear bilaterally towards the back of the head lie directly over the auditory cortex. 
These electrodes may be picking up on brain activation from the auditory cortex that is modulated by the perception of the stimuli. 
The electrodes within the blue areas that appear more centrally may be picking up on the cognitive processes that occur as a result of the brain processing the music at a higher level.
In layer 3 there is remarkable similarity of the learned temporal patterns between stimuli 1--4 and their corresponding stimuli 11--14, which are tempo-matched recordings of songs 1--4 without lyrics (cf.\,\autoref{tab:stimuli_information} for details).
This indicates that the models have picked up musical features from the stimuli such as down-beats (marking the beginning of a measure).
Further investigations will look at what aspects of the music are causing these similarities.

Ultimately, we would like to apply the EEG analysis techniques described here in a \acf{BCI} setting.
To this end, it would be already sufficient to distinguish between two stimuli with a high accuracy.
An analysis of the binary classification performance of model W showed already promising results with up to perfect accuracy for certain stimulus pairs
even though the model had been only trained for the 12-class problem (cf. \autoref{fig:model_W_binary_confusion}).

\section{Conclusions}

We have proposed several novel techniques for deep feature learning from EEG recordings that address specific challenges of this application domain.
Cross-trial encoders aim to capture invariance between trials within and across subjects.
Hydra-nets can learn specifically adapted versions of selected layers and switch between them based on trial metadata to, for instance, compensate for differences between subjects.
Finally, similarity-constraint encoders allow us to identify signal components that help to distinguish trials of the same class from others.
First experiments on the OpenMIIR dataset demonstrate that our techniques are able to learn features that are useful for classification.
Most significant improvements over a baseline using raw EEG input were obtained by similarity-constraint encoding picking up relevant brain regions as well musically meaningful temporal signal characteristics such as the position of down-beats.
Furthermore, the learned features are also still simple enough to allow interpretation by domain experts such as cognitive neuroscientists.

Whilst targeting problems specific to EEG analysis, the proposed techniques are not limited to this kind of data or even \acp{CNN}. 
Thus, we believe that they will be useful beyond the scope of our application domain.
We also hope that this paper will encourage other researchers to join us in the challenge of decoding brain signals with deep learning techniques.
For future work, we want to further refine our approach and investigate whether it can be used to identify overlapping features between music perception and imagery, which would be very useful for \ac{BCI} settings.

\subsubsection*{Acknowledgments}

This work has been supported by a fellowship within the Postdoc-Program of the German Academic Exchange Service (DAAD), the Canada Excellence Research Chairs (CERC) Program, an National Sciences and Engineering Research Council (NSERC) Discovery Grant, an Ontario Early Researcher Award, and the James S.\,McDonnell Foundation.
The authors would further like to thank the study participants.

\clearpage
\footnotesize{

\bibliographystyle{iclr2016_conference}
}

\clearpage
\appendix

\section{Additional Information about the OpenMIIR Music Stimuli}

\begin{table}[h]
\setlength{\tabcolsep}{3.5pt}  
\renewcommand{\arraystretch}{1}
\centering
\caption{Information about the tempo, meter and length of the stimuli (without cue clicks).}
\label{tab:stimuli_information}
{\footnotesize 
\begin{tabular}{r l c c  c c c }
\toprule
ID		& Name											&Meter 	& Length			& Tempo \\
\midrule
1		& Chim Chim Cheree (lyrics)				& 3/4		& 13.3s 			& 212 BPM \\
2		& Take Me Out to the Ballgame (lyrics)	& 3/4		& 7.7s 			& 189 BPM \\
3		& Jingle Bells (lyrics)							& 4/4		& 9.7s 			& 200 BPM \\
4		& Mary Had a Little Lamb (lyrics)			& 4/4		& 11.6s			& 160 BPM \\
11		& Chim Chim Cheree 							& 3/4		& 13.5s	 		& 212 BPM \\
12		& Take Me Out to the Ballgame			& 3/4		& 7.7s 			& 189 BPM \\
13		& Jingle Bells 									& 4/4		& 9.0s 			& 200 BPM \\
14		& Mary Had a Little Lamb					& 4/4		& 12.2s			& 160 BPM \\
21		& Emperor Waltz								& 3/4		& 8.3s 			& 178 BPM \\
22		& Hedwig's Theme (Harry Potter)			& 3/4		& 16.0s	 		& 166 BPM \\
23		& Imperial March (Star Wars Theme)		& 4/4		& 9.2s 			& 104 BPM \\
24		& Eine Kleine Nachtmusik					& 4/4		& 6.9s			& 140 BPM \\
\midrule
		& mean											&			& 10.4s			& 176 BPM \\ 
\bottomrule
\end{tabular}
}
\vspace{-1em}
\end{table}

\clearpage
\section{Visualizations of Common Components}

\begin{figure}[h] 
  \begin{center}
    \includegraphics[width=\textwidth,keepaspectratio=true]{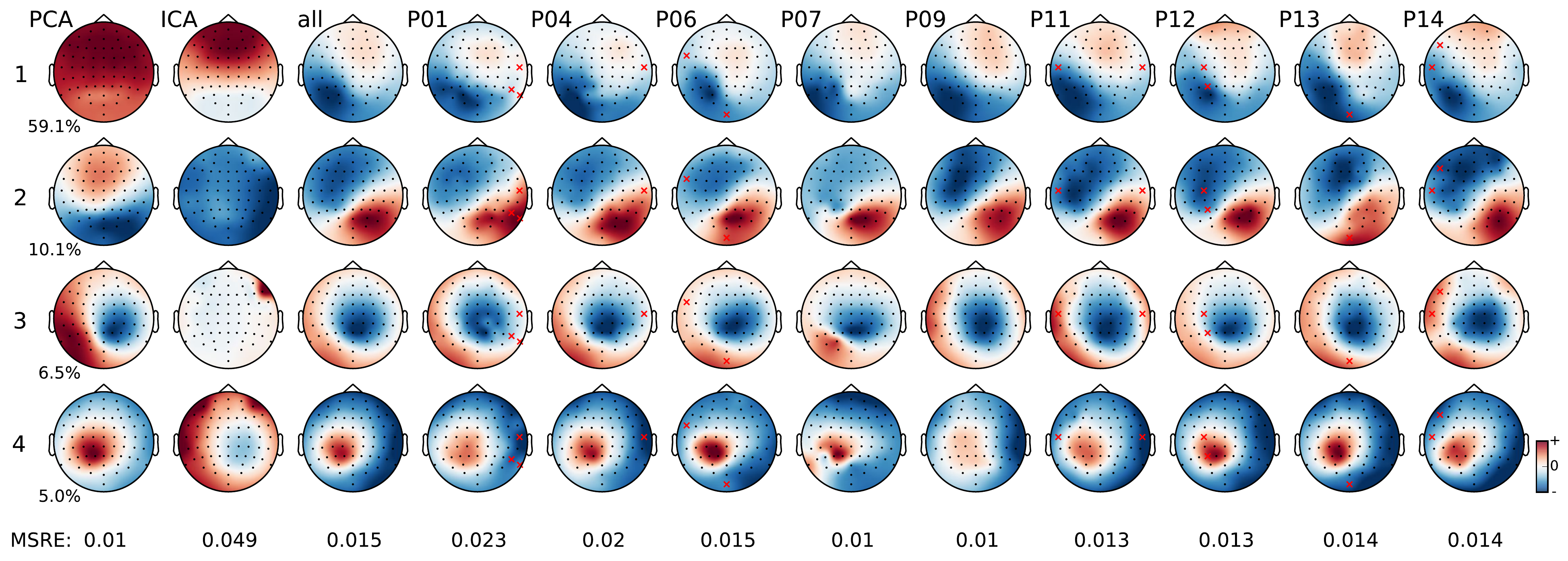}
    \caption{%
Topographic map visualization of the EEG channel weights for the top-4 signal components derived from the perception trials.
The leftmost two columns show the top-4 principal components (explaining over 80\% of the variance in total; individual values are shown next to the components) and the corresponding independent components computed using extended Infomax ICA.
The remaining columns show the weights learned by a tied-weights convolutional auto-encoder for all subjects combined (third column) 
and all individual subjects labeled P01 to P14 accordingly using 4 filters of width 1 (samples) with tanh nonlinearity and without biases.
Noisy EEG channels that were removed and interpolated are marked by red X's.
Values at the bottom of each column refer to the mean squared reconstruction error (MRSE) per sample, which was used as cost function.
}
    \label{fig:topoplots}
  \end{center}
\end{figure}

\begin{figure}[h] 
  \begin{center}
    \includegraphics[width=\textwidth,keepaspectratio=true]{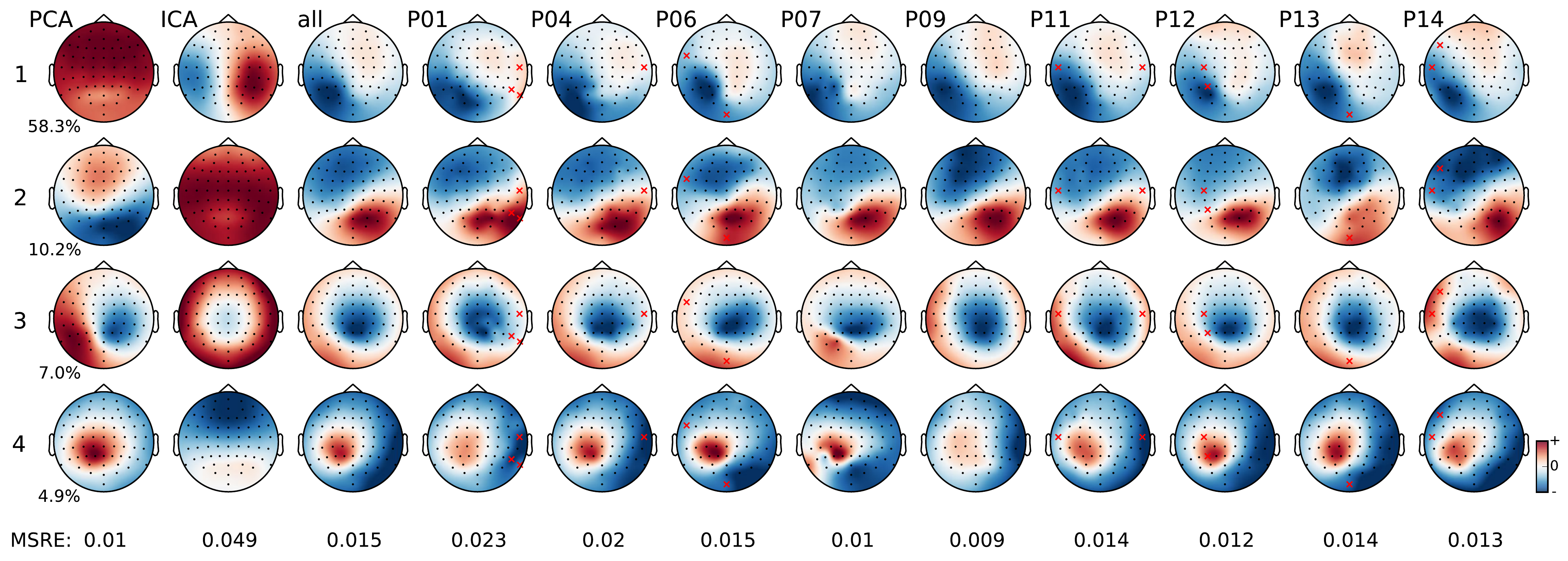}
    \caption{%
Topographic map visualization of the EEG channel weights for the top-4 signal components derived from the cued imagination trials.
They look very similar to the ones shown in \autoref{fig:topoplots} for the perception trials.
The leftmost two columns show the top-4 principal components (explaining over 80\% of the variance in total; individual values are shown next to the components) and the corresponding independent components computed using extended Infomax ICA.
The remaining columns show the weights learned by a tied-weights convolutional auto-encoder for all subjects combined (third column) 
and all individual subjects labeled P01 to P14 accordingly using 4 filters of width 1 (samples) with tanh nonlinearity and without biases.
Noisy EEG channels that were removed and interpolated are marked by red X's.
Values at the bottom of each column refer to the mean squared reconstruction error (MRSE) per sample, which was used as cost function.
}
    \label{fig:topoplots_imag}
  \end{center}
\end{figure}

\clearpage

\section{Further Visualizations of Cross-Trial Encoders}
\label{sec:CTE_more}

\begin{figure}[h] 
  \begin{center}
    \includegraphics[width=\textwidth,keepaspectratio=true]{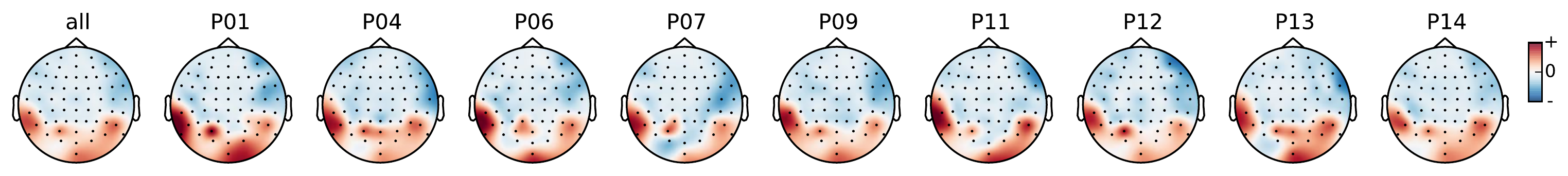}
    \caption{%
A single spatial filter after pre-training using cross-trial encoding.
Left column: Global filter after training on within-subject trial pairs (stage 1). 
Remaining columns: Individual filters (hydra-net) after training with cross-subject trials (stage 2).
This filter was used as fixed first layer in the \acp{CNN} classifiers J and K listed in \autoref{tab:results} and \autoref{tab:result_details}.
}
    \label{fig:topoplots_cte_oc1kw1}
  \end{center}
\end{figure}

\begin{figure}[h] 
  \begin{center}
    \includegraphics[width=\textwidth,keepaspectratio=true]{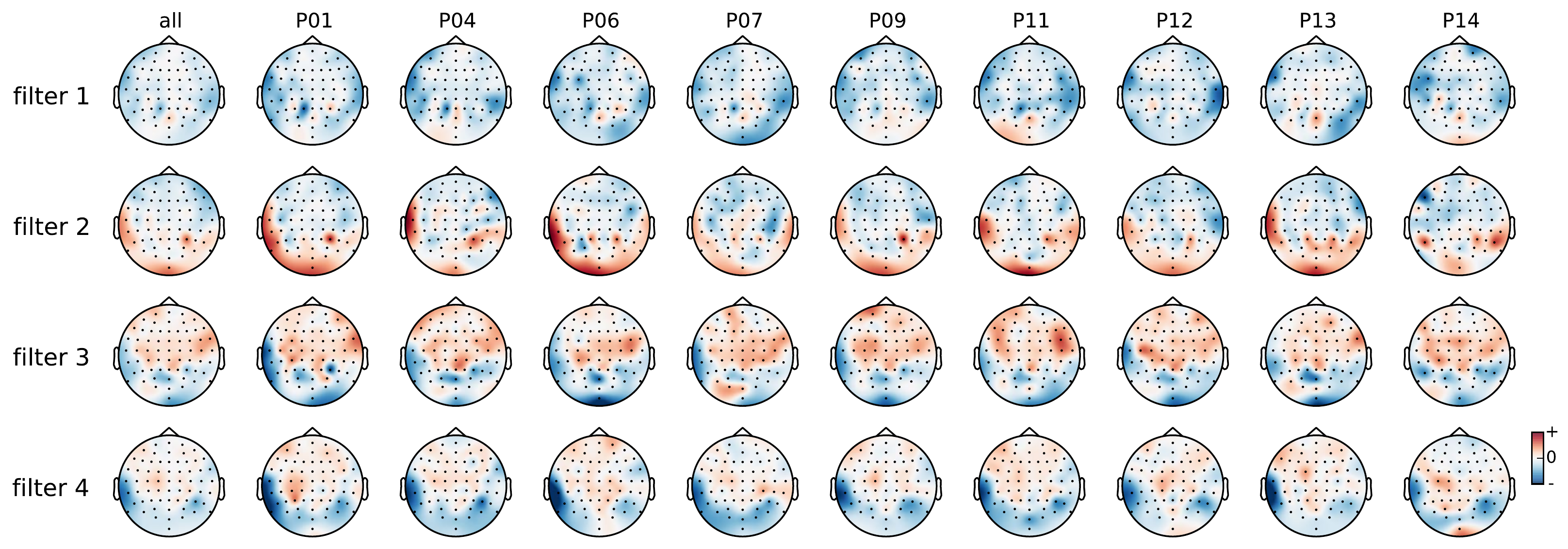}
    \caption{%
Four spatial filters after pre-training using cross-trial encoding.
Left column: Global filters after training on within-subject trial pairs (stage 1). 
Remaining columns: Individual filters (hydra-net) after training with cross-subject trials (stage 2).
This filter was used as fixed first layer in the \acp{CNN} classifiers N and O listed in \autoref{tab:results} and \autoref{tab:result_details}.
}
    \label{fig:topoplots_cte_oc4kw1}
  \end{center}
\end{figure}

\clearpage
\section{Detailed CNN Classifier Performance}\label{sec:detailed_results}
\begin{table}[h!]
\setlength{\tabcolsep}{2.25pt}  
\renewcommand{\arraystretch}{1.1}
\centering
\caption{Overview of best classifier \acp{CNN} identified by hyper-parameter optimization.
Layer parameters refer to output\_channels x [width x height] x input\_channels / sub-sampling\_factor.
}
\label{tab:result_details}
{\footnotesize  
\begin{tabular}{c l  c c c  c c c  r}
\toprule
ID	& description & valid.	& test (avg)	& test (maj) & 1st layer 	& 2nd layer & 3rd layer & \#params \\
\midrule 
\tablesec{Baseline (64\,Hz, no pre-training, 300 jobs for optimization):}
A & best test value 		& 33.3\% & 33.3\% & 27.8\% & 4x[6x1]x64 & 4x[9x1]x4 & 1708x12+12 & 22188 \\ 
B & best validation value & 39.6\% & 18.5\% & 26.9\% & 1x[5x1]x64 & 4x[10x1]x1 & 1708x12+12 & 20868 \\ 
C & smallest amongst best	& 35.4\% & 27.7\% & 25.9\% & 4x[3x1]x64 & 2x[12x1]x4 & 854x12+12	 & 11124 \\ 
\midrule  
\tablesec{Baseline (512\,Hz, no pre-training):}
D & best test value & 29.2\% & 23.1\% & 29.6\% & 3x[1x1]x64 & 1x[256x1]x3/7 & 466x12+12 & 6564 \\ 
E & best validation value & 31.2\% & 18.5\% & 19.4\% & 3x[1x1]x64 & 1x[256x1]x3/19 & 171x12+12 & 3024 \\ 
\midrule 
\tablesec{Common Components (64\,Hz, global filters):}
F & best test value & 29.2\% & 27.8\% & 25.9\% & 4x[1x1]x64 & 4x[16x1]x4 & 1700x12+12 & 20924 \\ 
G & best validation value & 33.3\% & 17.6\% & 18.5\% & 4x[1x1]x64 & 1x[16x1]x4 & 425x12+12 & 5432 \\ 
\midrule 
\tablesec{Common Components (64\,Hz, individually adapted filters):}
H & best test value & 29.2\% & 28.7\% & 25.0\% & 4x[1x1]x64 & 4x[16x1]x4 & 1700x12+12 & 20924 \\ 
I & best validation value & 31.2\% & 20.4\% & 20.4\% & 4x[1x1]x64 & 8x[3x1]x4 & 3504x12+12 & 42412 \\ 
\midrule 
\tablesec{Cross-Trial Encoder (64\,Hz, individually adapted filters):}
J & best test value & 33.3\% & 25.0\% & 23.1\% & 1x[1x1]x64 & 16x[7x1]x1 & 6944x12+12 & 83516 \\ 
K & best validation value & 35.4\% & 20.4\% & 19.4\% & 1x[1x1]x64 & 1x[6x1]x1 & 435x12+12 & 5302 \\ 
L & best test value & 31.2\% & 25.0\% & 24.1\% & 1x[3x1]x64 & 1x[1x1]x1 & 438x12+12 & 5461 \\ 
M & best validation value & 39.6\% & 23.1\% & 22.2\% & 1x[3x1]x64 & 1x[4x1]x1 & 435x12+12 & 5428 \\ 
N & best test value & 31.2\% & 27.8\% & 25.0\% & 4x[1x1]x64 & 14x[1x1]x4/3 & 2044x12+12 & 24852 \\ 
O & best validation value & 39.6\% & 22.2\% & 22.2\% & 4x[1x1]x64 & 10x[1x1]x4 & 4400x12+12 & 53108 \\ 
\midrule 
\tablesec{Similarity-Constraint Encoder (64\,Hz):}
P & best test value & 33.3\% & 28.7\% & 25.9\% & 1x[1x1]x64 & 14x[9x1]x1/2 & 3024x12+12 & 36490 \\ 
Q & best validation value & 35.4\% & 25.0\% & 27.8\% & 1x[1x1]x64 & 10x[11x1]x1/2 & 2150x12+12 & 25986 \\ 
R & smallest amongst best & 33.3\% & 25.9\% & 25.9\% & 1x[1x1]x64 & 1x[1x1]x1 & 440x12+12 & 5346 \\ 
S & best test value & 31.2\% & 26.9\% & 27.8\% & 1x[3x1]x64 & 16x[5x1]x1 & 6944x12+12 & 83612 \\ 
T & best validation value & 35.4\% & 20.4\% & 18.5\% & 1x[3x1]x64 & 10x[10x1]x1/3 & 1430x12+12 & 17464 \\ 
\midrule 
\tablesec{Similarity-Constraint Encoder (512\,Hz):}
U & best test value & 29.2\% & 33.3\% & 34.3\% & 1x[1x1]x64 & 13x[85x1]x1/10 & 4459x12+12 & 54689 \\ 
V & best validation value & 35.4\% & 22.2\% & 23.1\% & 1x[1x1]x64 & 7x[69x1]x1/14 & 1722x12+12 & 21223 \\ 
W & smallest amongst best & 33.3\% & 28.7\% & 27.8\% & 1x[1x1]x64 & 1x[37x1]x1/11 & 316x12+12 & 3905 \\ 
X & best test value & 29.2\% & 31.5\% & 31.5\% & 1x[3x1]x64 & 16x[128x1]x1 & 54224x12+12 & 652940 \\ 
Y & best validation value & 35.4\% & 27.8\% & 26.9\% & 1x[3x1]x64 & 5x[40x1]x1/7 & 2480x12+12 & 30164 \\ 
\bottomrule
\end{tabular}
}
\vspace{-1em}
\end{table}

\clearpage
\section{Model Visualizations}\label{sec:model_visualizations}

\begin{figure}[h] 
  \begin{center}
    \includegraphics[width=\textwidth,keepaspectratio=true]{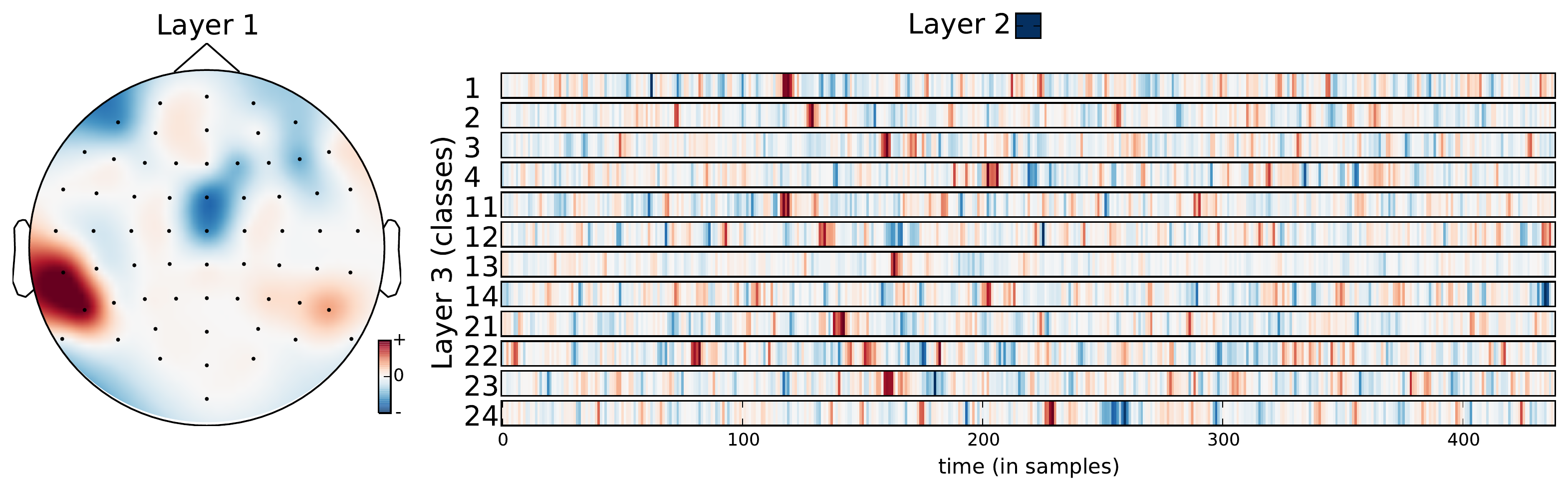}
    \caption{Visualization of \ac{CNN} R (average of the 9 cross-validation fold models), which processes raw EEG at a sampling rate of 64\,Hz.
    Layer 1 was pre-trained using similarity-constraint encoding.}
    \label{fig:model_R}
  \end{center}
\end{figure}

\begin{figure}[h] 
  \begin{center}
    \includegraphics[width=\textwidth,keepaspectratio=true]{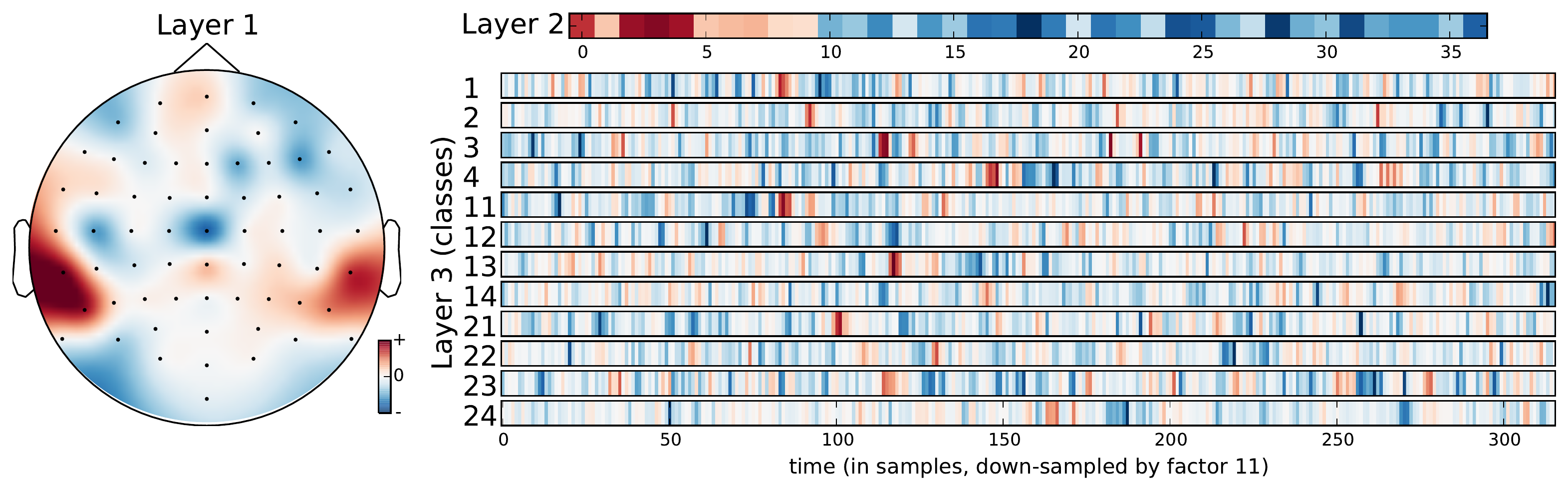}
    \caption{Visualization of \ac{CNN} W (average of the 9 cross-validation fold models), which processes raw EEG at a sampling rate of 512\,Hz.
    Layer 1 was pre-trained using similarity-constraint encoding.}
    \label{fig:model_W}
  \end{center}
\end{figure}

\clearpage
\section{Confusion Analysis for CNN Model W (avg)}\label{sec:confusion}

\vspace{-1em}
\begin{figure}[h] 
  \begin{center}
    \includegraphics[width=.65\textwidth,keepaspectratio=true]{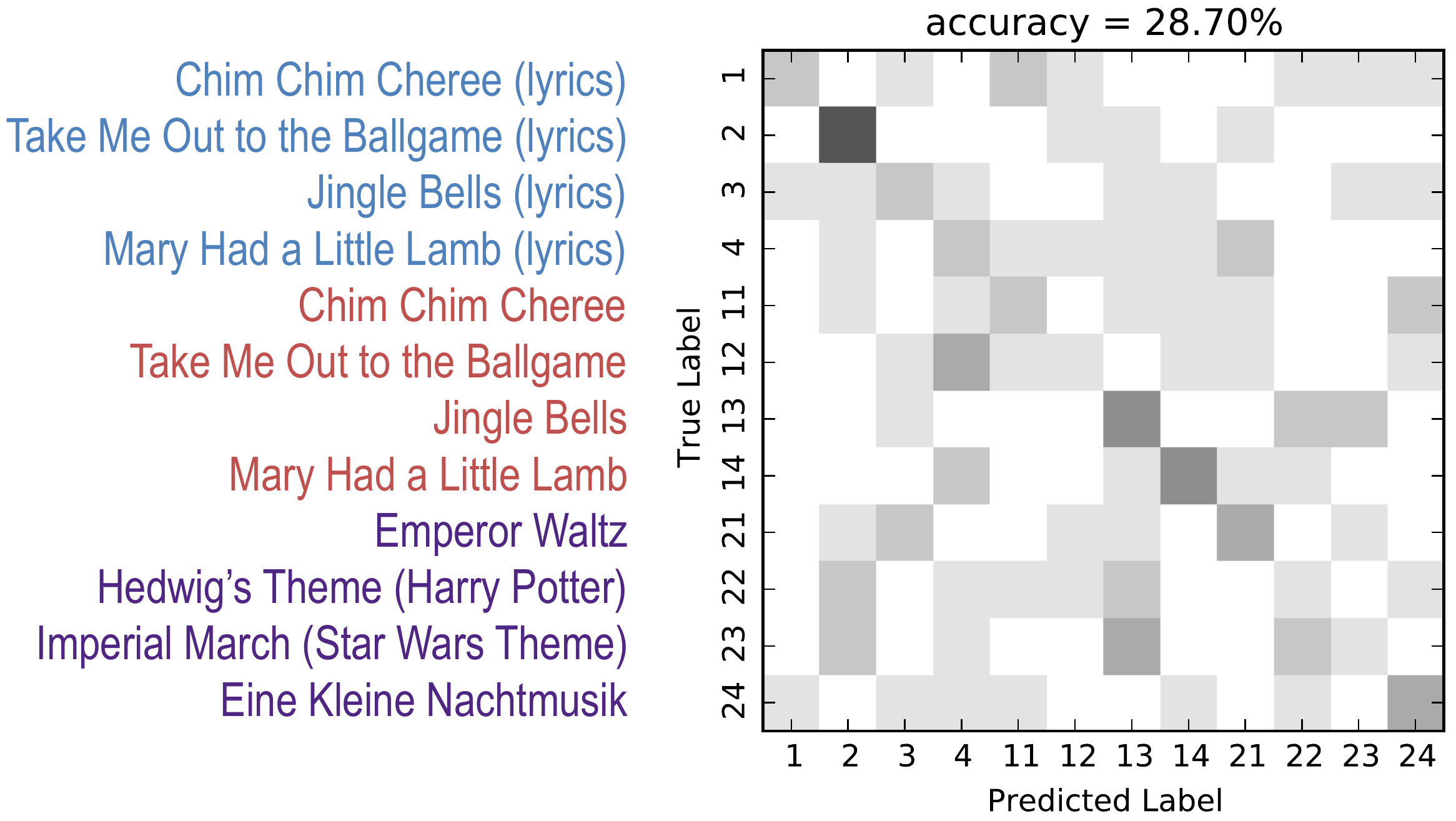}
   \\\vspace{-0.8em}
    \caption{12-class confusion matrix for \ac{CNN} W (average of the 9 cross-validation fold models).}
    \label{fig:model_W_confusion}
  \end{center}
  \vspace{-1em}
\end{figure}

\begin{figure}[h] 
  \begin{center}
    \includegraphics[width=.93\textwidth,keepaspectratio=true]{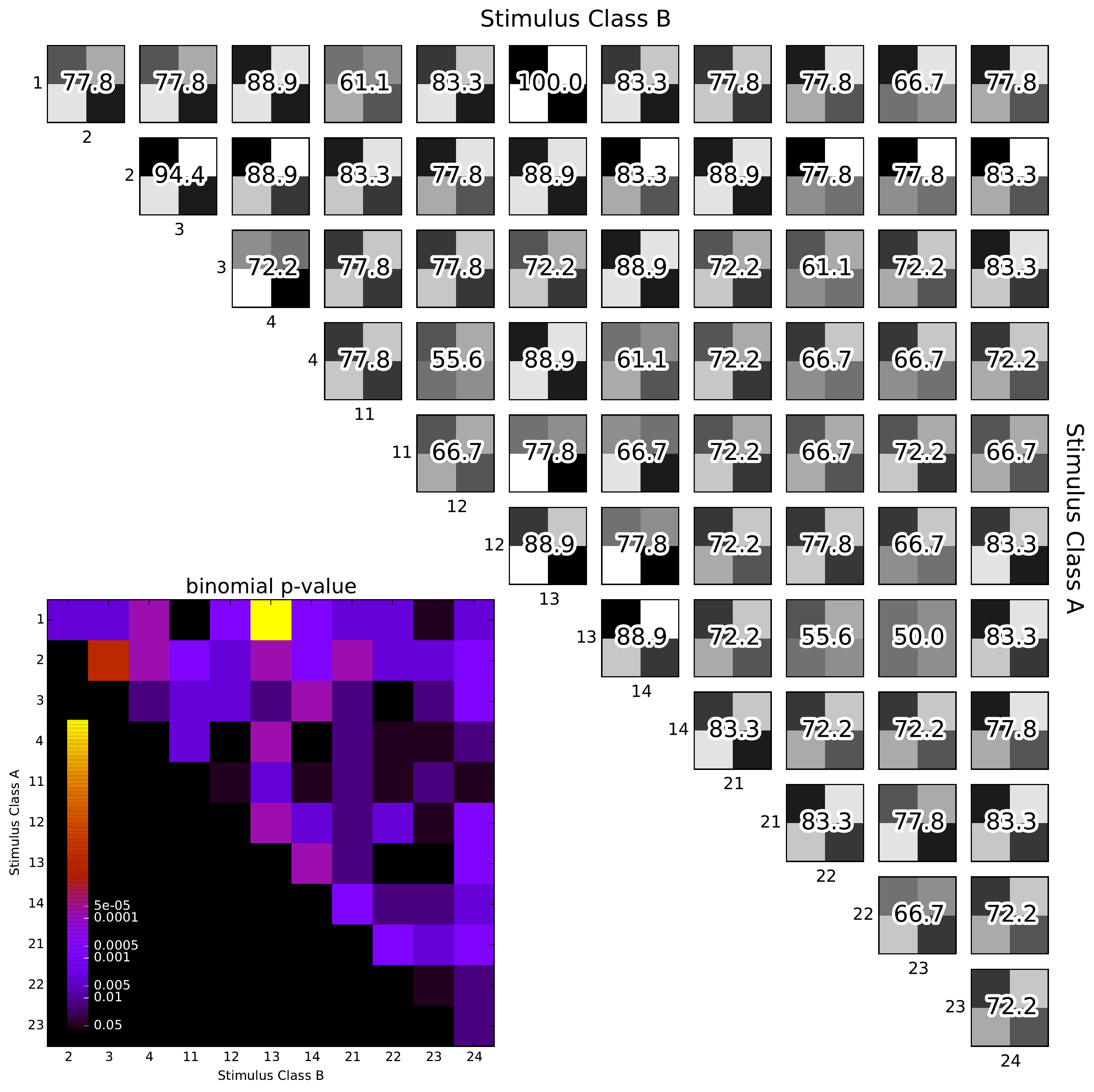}
   \\\vspace{-0.8em}
    \caption{Binary confusion matrices for \ac{CNN} W (average of the 9 cross-validation fold models).
    For each binary classification, only the 18 test trials belonging to either stimulus class A or stimulus class B were considered.
    The inset at the lower bottom visualizes the p-values determined by using the cumulative binomial distribution to estimate the likelihood of observing the resepective binary classification rate by chance.
    Note that the classifier was only trained for the 12-class problem.}
    \label{fig:model_W_binary_confusion}
  \end{center}
  \vspace{-1em}
\end{figure}

\clearpage
\section{Implementation Details}\label{sec:implementation}

For 
reproducibility and to encourage further developments and research in this direction,
all code necessary to build the proposed deep network structures and to run the experiments described in the following is shared as open source within the \emph{deepthought} library.\footnote{%
\url{https://github.com/sstober/deepthought} (code will be updated paper publication)
}	
The implementation is based on the libraries \emph{Pylearn2} (\cite{goodfellow_pylearn2:_2013}) and \emph{Theano} (\cite{Bergstra2010Theano}) and comprises various custom \emph{Layer} and \emph{Dataset} classes --
such as for on-the-fly generation of trial tuples and the respective classification targets during iteration.

The following subsections provide details about our implementation.

\subsection{Details on Convolutional Auto-Encoders for EEG}

Convolutional autoencoders are a special variant of \acp{CNN} that encode their input using convolution into a compressed internal representation which is then decoded using de-convolution into the original space trying to minimize the reconstruction error.
Such encoder/decoder pairs can optionally be stacked as described in \cite{masci_stacked_2011} to increase the complexity of the internal representations.

We considered two measures of the reconstruction quality.
The \ac{MSRE} is the commonly used error measure for auto-encoding and computed as the squared Euclidean distance between the input and the reconstruction averaged over all samples.
We further define the \ac{MCC} as the mean correlation of the individual input channels and their reconstructions.
The correlation is computed as Pearson's r, which is in the range [-1,1] where 1 is total positive correlation, 0 is no correlation, and -1 is total negative correlation.

The convolutional filters applied on multi-channel raw EEG data are 2-dimensional where the width is time (samples) and the height is optionally used for different frequency bands in a time-frequency representation, which is not considered in the context of this paper.
In the simplest case, the filter width in the time dimension is only 1.
A \acp{CNN} with $k$ filters then behaves like a conventional feed-forward network without convolution that aggregates \ac{EEG} data for a single time point across all channels into $k$ output values -- with the only difference that the \ac{CNN} processes the whole recording as one input.

Such a convolutional auto-encoder with filter width 1 and linear activation functions is strongly related to \ac{PCA} (\cite{bourlard_auto-association_1988}) and consequently obtains similar results.
Using an auto-encoder, however, opens up further possibilities such as non-linear activation functions, a filter width greater than 1, regularization of channel weights and activations, and stacking multiple encoder/decoder layer pairs.
This allows to identify more complex components, which can also cover the time dimension.

Here, we used all training trials from the dataset for training which were preprocessed as described in \autoref{sec:dataset}.
The training trials have different lengths due to the different duration of the stimuli.
As the network expects inputs of equal length, shorter trials were zero-padded at the end to match the duration of the longest stimulus.\footnote{%
This is a limitation of our current \emph{Pylearn2}-based implementation.
In principle, convolutional auto-encoders can operate on variable-length input.
}
Zero-padded trials should not be used for supervised training because this will result in the trial length being used as a distinctive feature, which is not desirable.
However, for unsupervised pre-training using basic auto-encoding, the variable trial length is not problematic as the objective here is only to reconstruct.

The de-convolution filters were set to share the weights with the corresponding convolution filters.
We experimented with different activation functions -- linear, tanh and rectified linear (\cite{Glorot2010Deep-Sparse-Rectifier}) -- and obtained very similar topographies that only had small variations in intensity.
We finally selected the tanh nonlinearity because its output matches the value range of the network inputs ([-1,1]).
Furthermore, we found that using bias terms did not lead to improvements and therefore chose network units without bias.

Comparing the commonly used \ac{MSRE} to the \ac{MCC} measure showed that \ac{MSRE} was the superior cost function in this setting. 
The training process was generally very stable such that a high learning rate with linear decay over epochs in combination with momentum could be used to quickly obtain results.
As termination criterion, we tracked the error on the training set for early stopping and specified a (rather defensive) maximum of 1000 epochs.

We first trained an auto-encoder for all trials of the perception condition.
The learned filter weights were then used as initialization values for individual auto-encoders for each subject.
Minor differences between subjects can be recognized in the example shown in \autoref{fig:topoplots} but the component relations between subjects are still obvious.
This was generally the case for different network structures that we tested.
The individual components seem to be very stable.
For instance, the results obtained with the same network on the imagination trials with cue (condition 2; otherwise not considered in the context of this paper) only differ slightly as shown in \autoref{fig:topoplots_imag}.

\subsection{Implementing a Hydra-Net}\label{sec:hydranet_details}

Hydra-nets allow to have separate processing pathways for subsets of a dataset 
-- such as all trials of the same subject 
or all trials belonging to the same stimulus.
We call the respective meta-data property that controls the choice of the individual pathway the \emph{selector}.
The selector data needs to be provided additionally to the regular network input and, if necessary, has to be propagated (without being processed) up to the point where it is used to choose the individual processing pathway.
Individualization can range from a single layer with different weights for each value of the selector
up to multi-layer sub-networks that may also differ in their structure as long as their input and output format is identical.

In our implementation, we wrapped the part of the network that we wanted to individualize with a hydra-net layer.
Within this wrapper layer, a copy of the internal network was created for each possible selector value.
For each input, the respective individualized version of the internal network was chosen based on the additionally provided selector input.
For mini-batch processing, we implemented two different strategies:
\begin{enumerate}[label=\alph*)]
\item
	Compute the output of all individual pathways and apply a mask based on the selector values for the whole mini-batch.
\item
	Iterate through the input instances within the mini-batch (using Theano's \emph{scan} operation) and process each one individually solely with the selected individual pathway. (For selection, Theano's \emph{ifelse} operation was used that implements a lazy evaluation.)
\end{enumerate}
Option a) is highly inefficient as it results in an overhead of unnecessary computations.
This effect gets worse with an increasing number of different selector values.
For option b), this is not the case.
However, there is a different performance penalty from using the scan function.
In the experiments described in \autoref{sec:CTE}, we found that option a) still performed slightly better.
We plan to optimize our code and investigate different ways to implement hydra-net layers for future experiments.

\subsection{Training a Cross-trial Encoder with Cross-Subject Trial Pairs}

\begin{figure}[h] 
  \begin{center}
      \includegraphics[width=\textwidth,keepaspectratio=true]{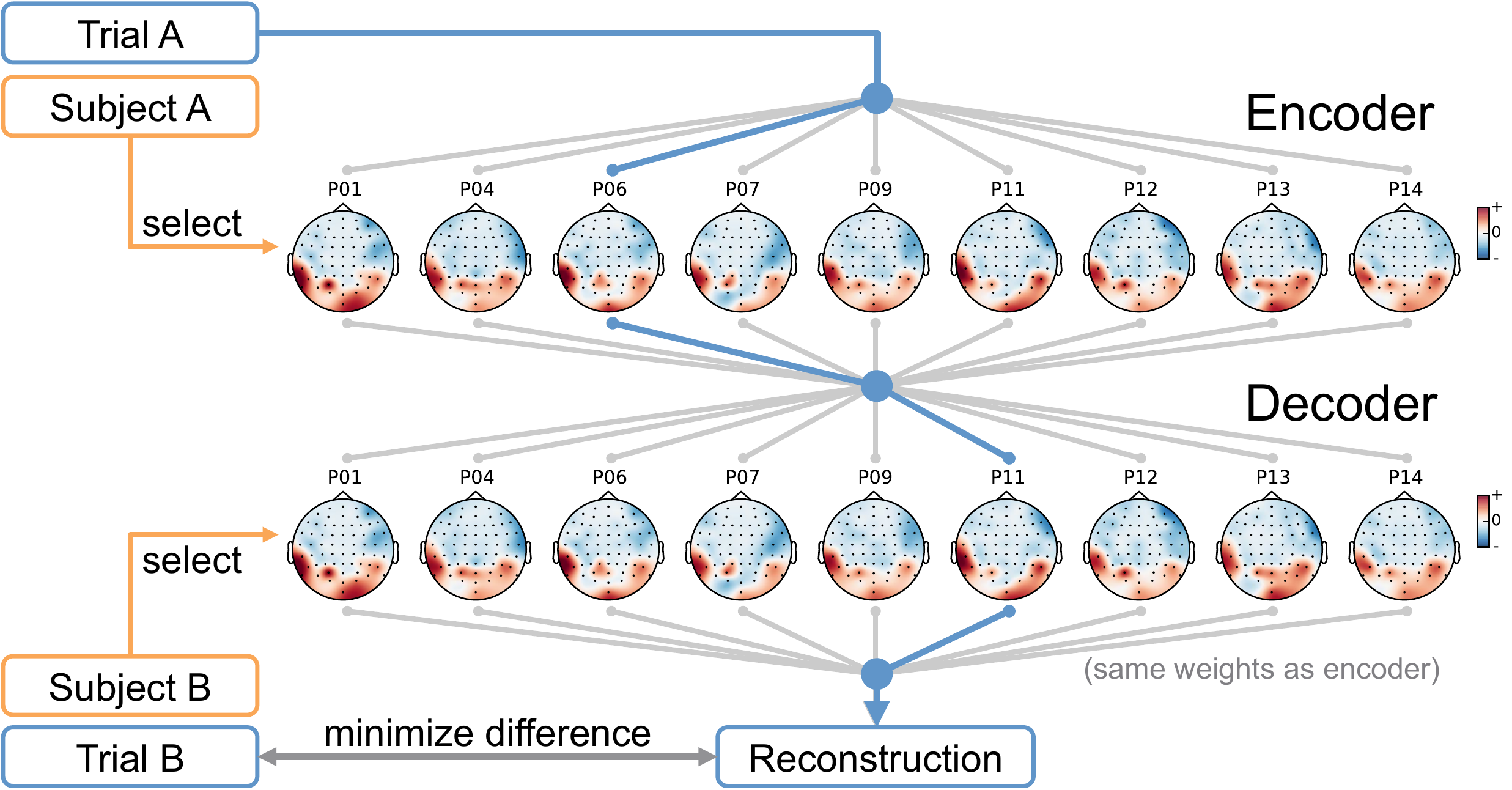}
    \caption{%
Cross-trial encoder training with cross-subject trial pairs (A,B). 
Both trials always belong to the same class.
Two separate hydra-net layers with shared weights (within subjects) were used as encoder (convolution) and decoder (deconvolution).
Encoder weights were selected based on the subject meta-data of the input trial (A)
whereas the decoder weights were selected based on the subject meta-data of the paired trial (B).
In this example, a trial from subject P06 is paired with a trial from subject P11.
The respective processing pathway for these selector values is highlighted in blue.
During backpropagation of the reconstruction error, only the weights of the filter weights along the selected pathway are adapted.
}
    \label{fig:CTE-hydra-net}
  \end{center}
\end{figure}

\autoref{fig:CTE-hydra-net} shows the network structure of the auto-encoder that learned the spatial filters shown in \autoref{fig:topoplots_cte_oc1kw1}.
Here, we used two separate hydra-net layers -- one as encoder and the other one as decoder.
Both shared their weights, i.e., for the same subject, the convolution and deconvolution filter would be identical.
The encoder hydra-net layer used the subject of the input trial as selector (subject A)
whereas the decoder hydra-net layer used the subject of the paired trial (subject B).
This general structure was identical for all cross-trial encoders considered here except for the shape and number of the filters used.

For initialization, we first trained a basic auto-encoder on all 1296 (= 432x3) within-subject trial pairs derived from the training set.
This way, we obtained the general filters labeled ``all'' in Figures \ref{fig:topoplots_cte_oc1kw3}, \ref{fig:topoplots_cte_oc1kw1} and \ref{fig:topoplots_cte_oc4kw1}.
We then continued trained a copy of the general cross-trial encoder for each individual subject.
Finally, we used the resulting individual filters as initialization for the hydra-net.
This network was trained on 15120 (= 432x(4x9 - 1)) cross-subject trial pairs.
In order to avoid excessive memory usage of the dataset, we implemented a dataset wrapper that only stored an index data structure and generated mini-batches of paired trials and selector meta-data on-the-fly from a base dataset when they were requested.

\subsection{Constructing and Training a Similarity-Constraint Encoder}\label{sec:SCE_details}

\begin{figure}[thb] 
  \begin{center}
      \includegraphics[width=0.8\textwidth,keepaspectratio=true]{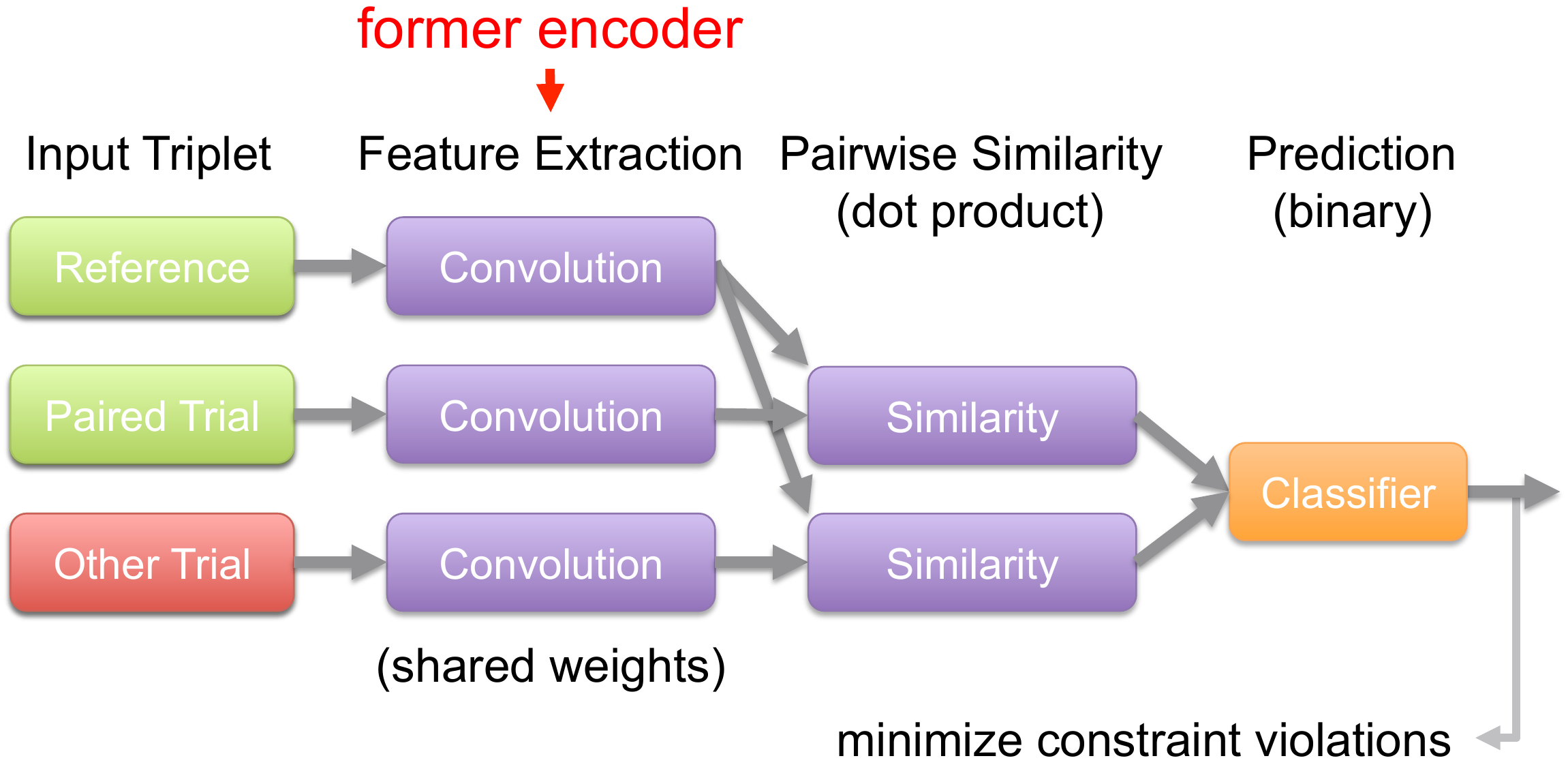}
    \caption{%
Processing scheme of a similarity-constraint encoder as used in the experiments described in this paper.
In general, the feature extraction part can also be more complex than shown here and does not necessarily involve convolution.
}
    \label{fig:SCE-scheme}
  \end{center}
\end{figure}

Starting from a basic auto-encoder, a similarity-constraint encoder as shown schematically in \autoref{fig:SCE-scheme} can be constructed as follows:
The decoder part of the original auto-encoder can be removed as we are now only interested in the internal feature representations and their similarity.\footnote{%
Note that simply maximizing the feature similarity for the paired trials used for cross-trial encoding would be an ill-posed learning problem.
The optimum could easily be achieved by transforming all inputs into the same representation.
This way, all trials would have maximum similarity -- including those from different classes.
It would also be impractical to train the network with pre-defined target similarity values for all possible pairs of trials.
}
Next, we redesign the encoder part such that it can process multiple trials at the same time as a single input tuple.
To ensure that all trials within a tuple are processed in the exactly same way, weights and biases need to be shared between the parallel processing pipelines.
One easy way to achieve such a weight sharing for basic \ac{CNN}-based architectures is by adding a third dimension for the trials (additionally to the channels and time/samples axes) and using filters of size 1 along this dimension.
Unfortunately, this approach of convolution along the trials axis does not work together with hydra-net layers.
Here, different individualized weights for the different trials of the input tuple might be required within the processing pipeline.
In our implementation, we therefore wrap the processing pipeline within a custom layer that takes care of processing each trial separately (using individualized weights if necessary) and then computes the pair-wise similarities between the reference trial and the other trials.
For $k$-tuples, this results in $k-1$ similarity scores.
The final output layer only has to predict the trial with the highest value.
This can, for instance, be accomplished with a Softmax (without affine transformation).
A variety of similarity measures could be used to compute the pair-wise similarities.
We choose the dot product because this matches the computation in the classifier output layer -- independent of whether it is a Softmax or hinge loss (DLSVM) layer. 
More complex approaches are possible as well, as long as they allow training through backpropagation -- such as first performing a channel-wise comparison and then aggregating the individual scores using trainable weights.

The dataset for training comprised 57024 within-subject trial triplets. 
This number results from pairing each of the 432 trials from the training set with the 3 other trials from the same class and subject
and further combining these pairs with the 11x4 trials from the same subject that belong to a different class.
As for cross-trial encoder training, we again implemented a dataset wrapper that only requires storing an index data structure for the tuples from a base dataset.

\subsection{Similarity-Constraint Encoders with Hydra-Net Layers}

We believe that the combination of similarity-constraint encoding with the hydra-net approach to account for individual differences between subjects has a high potential to further improve pre-trained global filters and the resulting classification accuracy.
An experiment with synthesized data has already shown the feasibility of this combined approach.
For this test, we generated random single-channel signals for 12 stimuli.
These signals were then added to 64-channel Gaussian noise, randomly choosing a different channel for each subject to contain the relevant signal.
As in our real dataset, we had 9 different subjects.
Using these synthetic data, a basic similarity-constraint encoder was able to identify the combined set of channels that contained relevant signals for the group of all subject.
Turning the first layer into a hydra-net layer allowed the spatial filters to become more specific and the network correctly learned the individual channel masks.

Training on the real dataset with less defined signals and much more structured background noise is of course much harder.
However, the biggest obstacle is currently the increase in processing time caused by the added computational complexity of the hydra-net layer (cf.\ \autoref{sec:hydranet_details}) with the amount of triplets available for training.
From the training set of 432 trials, 57024 within-subject trial triplets can be derived as described earlier.
Considering also cross-subject trial triplets increases the number to 5987520 (= 432x(4x9-1)x(11x4x9)).
Using training sets of this size is still feasible. 
However, the hydra-net processing overhead should be addressed before attempting this.

\clearpage
\section{Classifier Performance during Hyper-Parameter Optimization}\label{sec:scatterplots}

The following scatter plots show the estimated error (computed as the median over the 9-fold cross-validation sets) and the actual test set error obtained for the averaged model (left) and the majority vote from the 9 fold models (right) for all models trained during hyper-parameter optimization.
Selected models listed in \autoref{tab:results} are highlighted and labeled with their ID used in the table.

  \begin{center}
      \includegraphics[width=\textwidth,keepaspectratio=true]{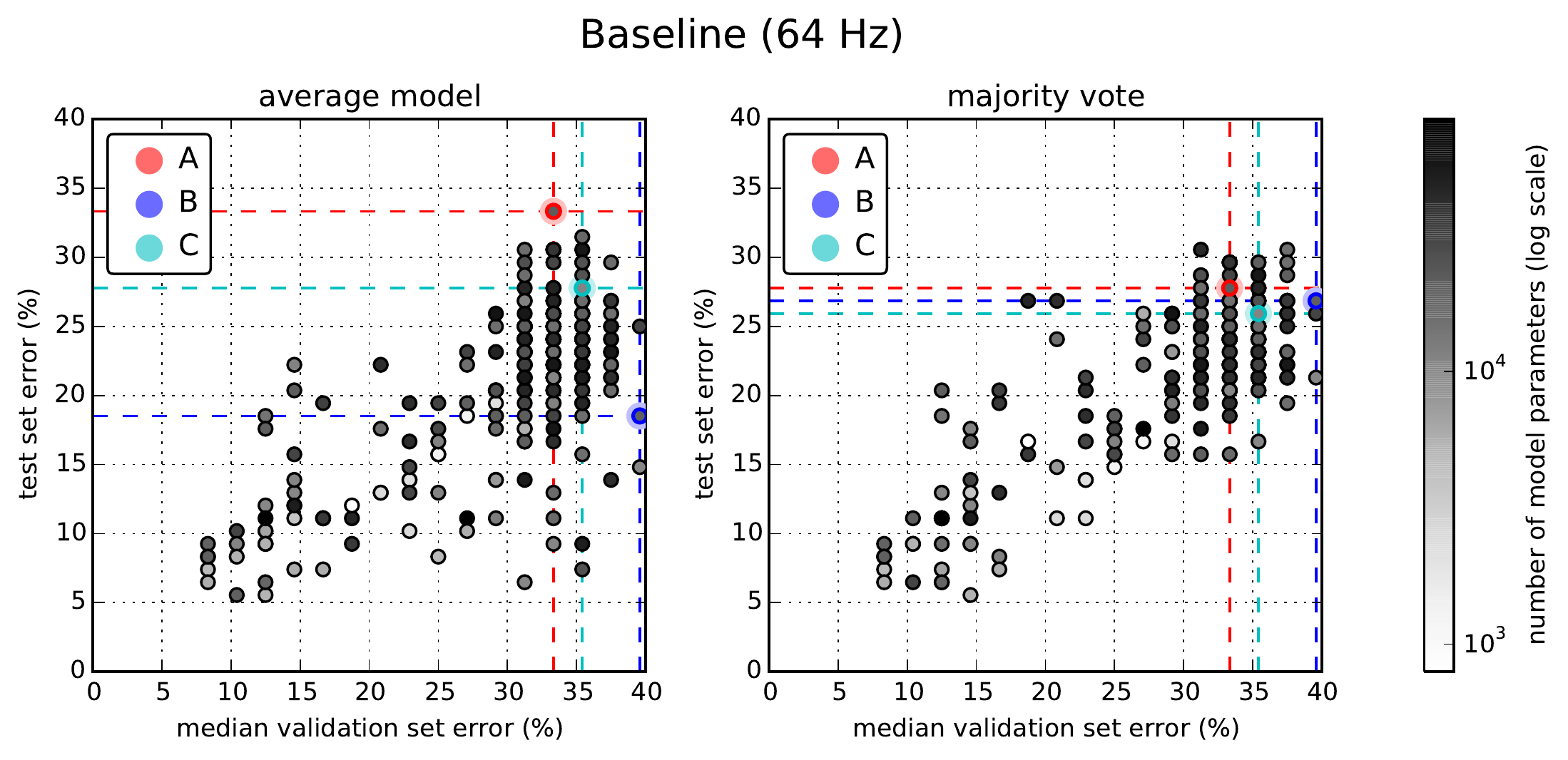}
      \includegraphics[width=\textwidth,keepaspectratio=true]{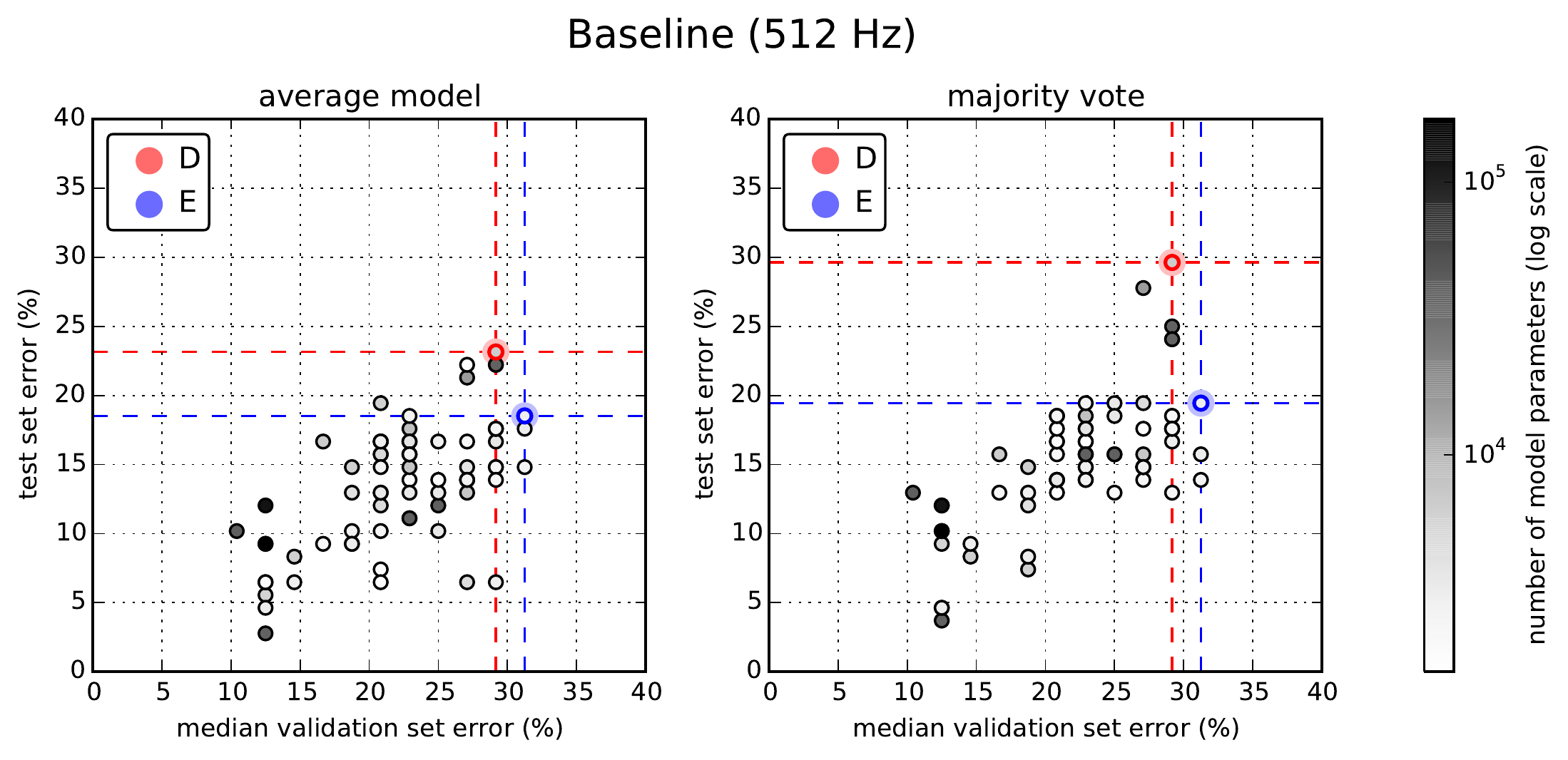}
  \end{center}
\clearpage

  \begin{center}
    \includegraphics[width=\textwidth,keepaspectratio=true]{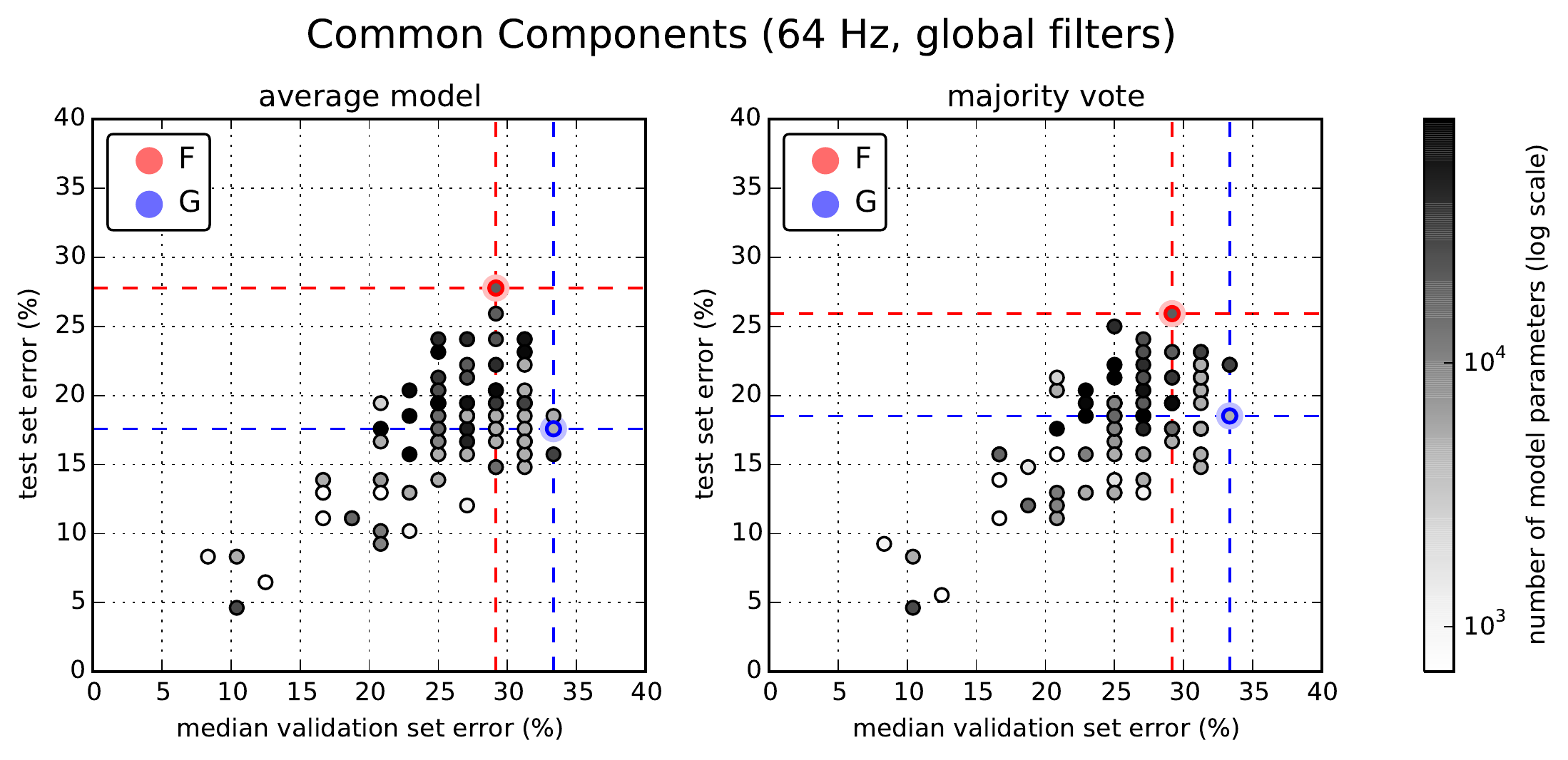}
    \includegraphics[width=\textwidth,keepaspectratio=true]{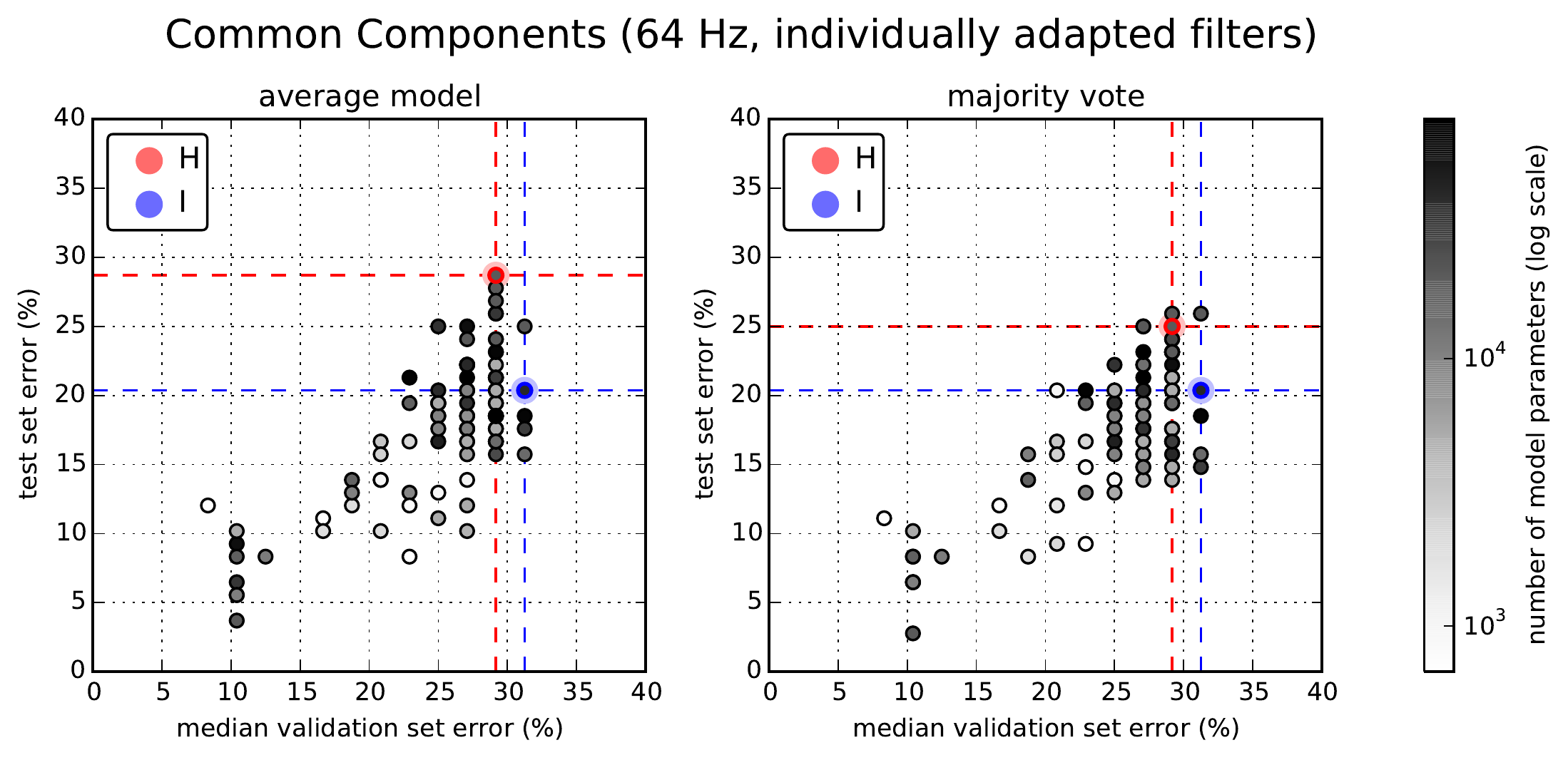}
  \end{center}
\clearpage

  \begin{center}
    \includegraphics[width=\textwidth,keepaspectratio=true]{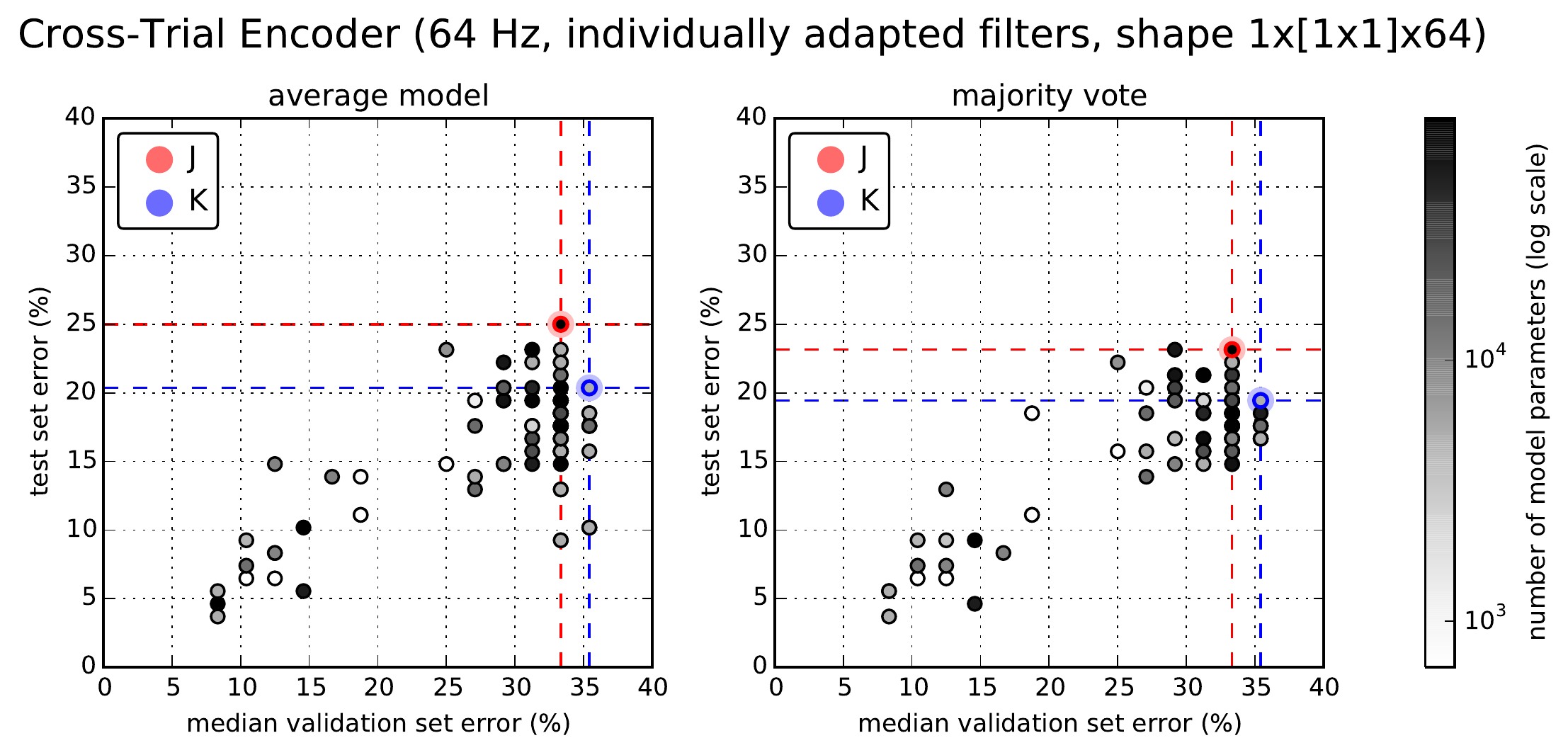}
    \includegraphics[width=\textwidth,keepaspectratio=true]{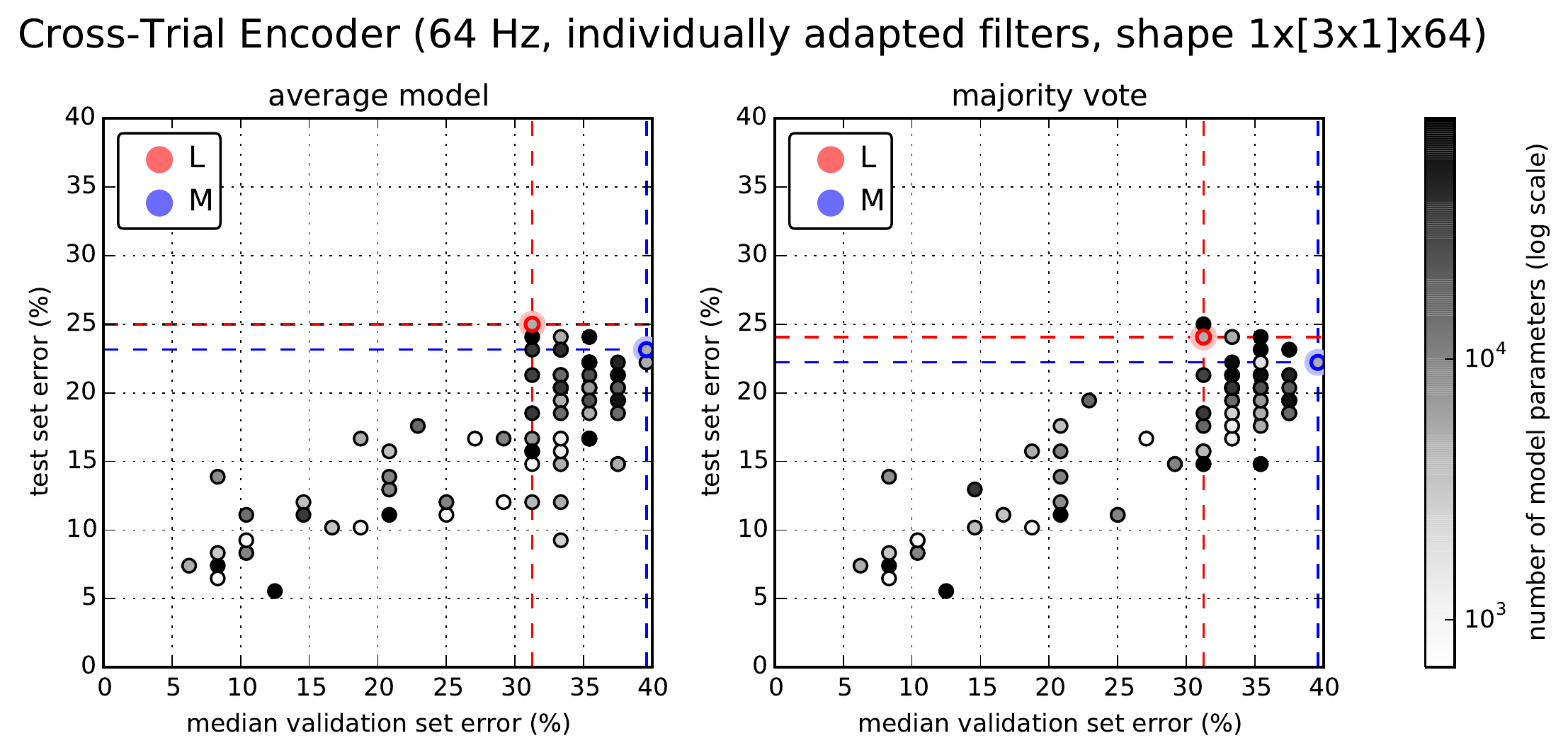}
    \includegraphics[width=\textwidth,keepaspectratio=true]{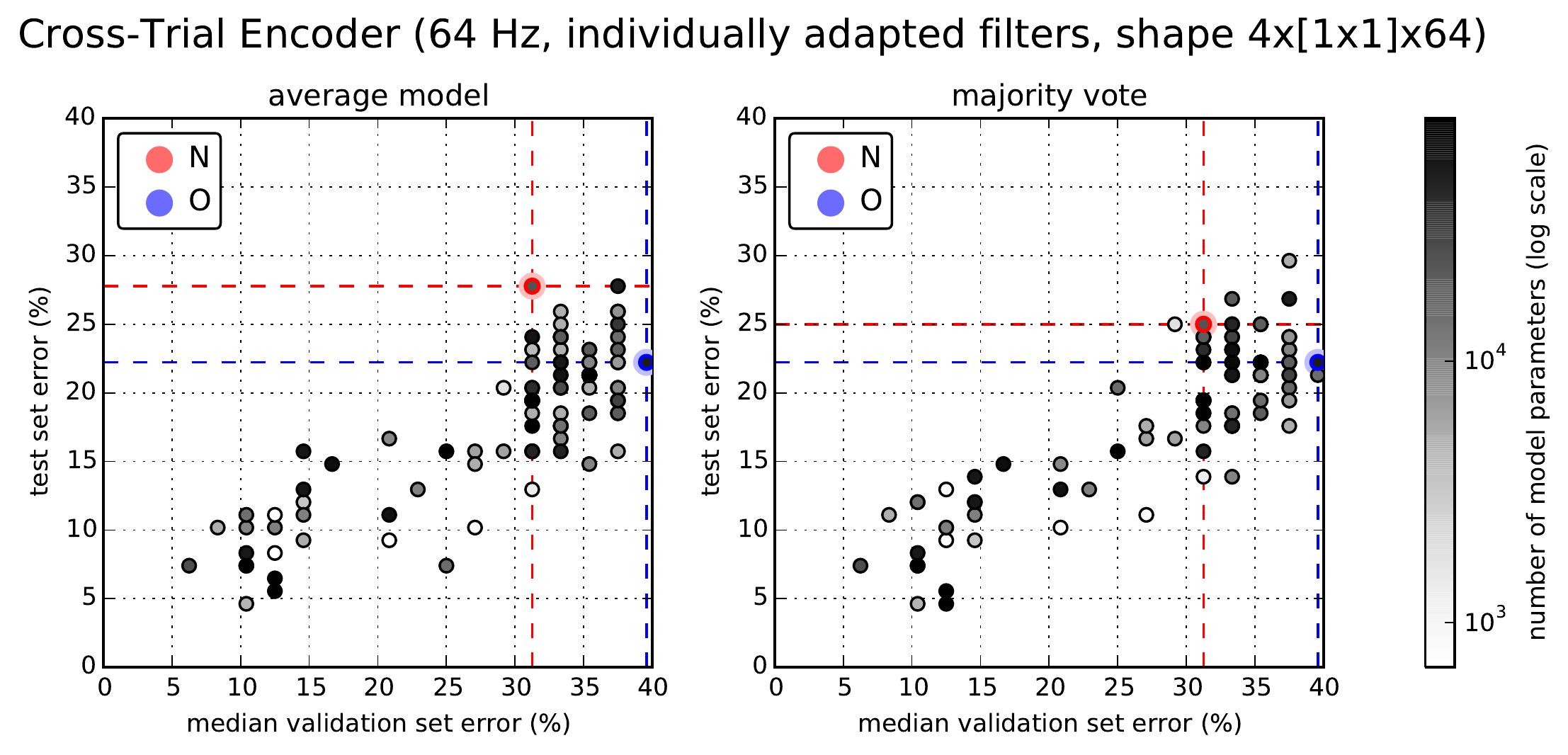}
  \end{center}
\clearpage

  \begin{center}
    \includegraphics[width=\textwidth,keepaspectratio=true]{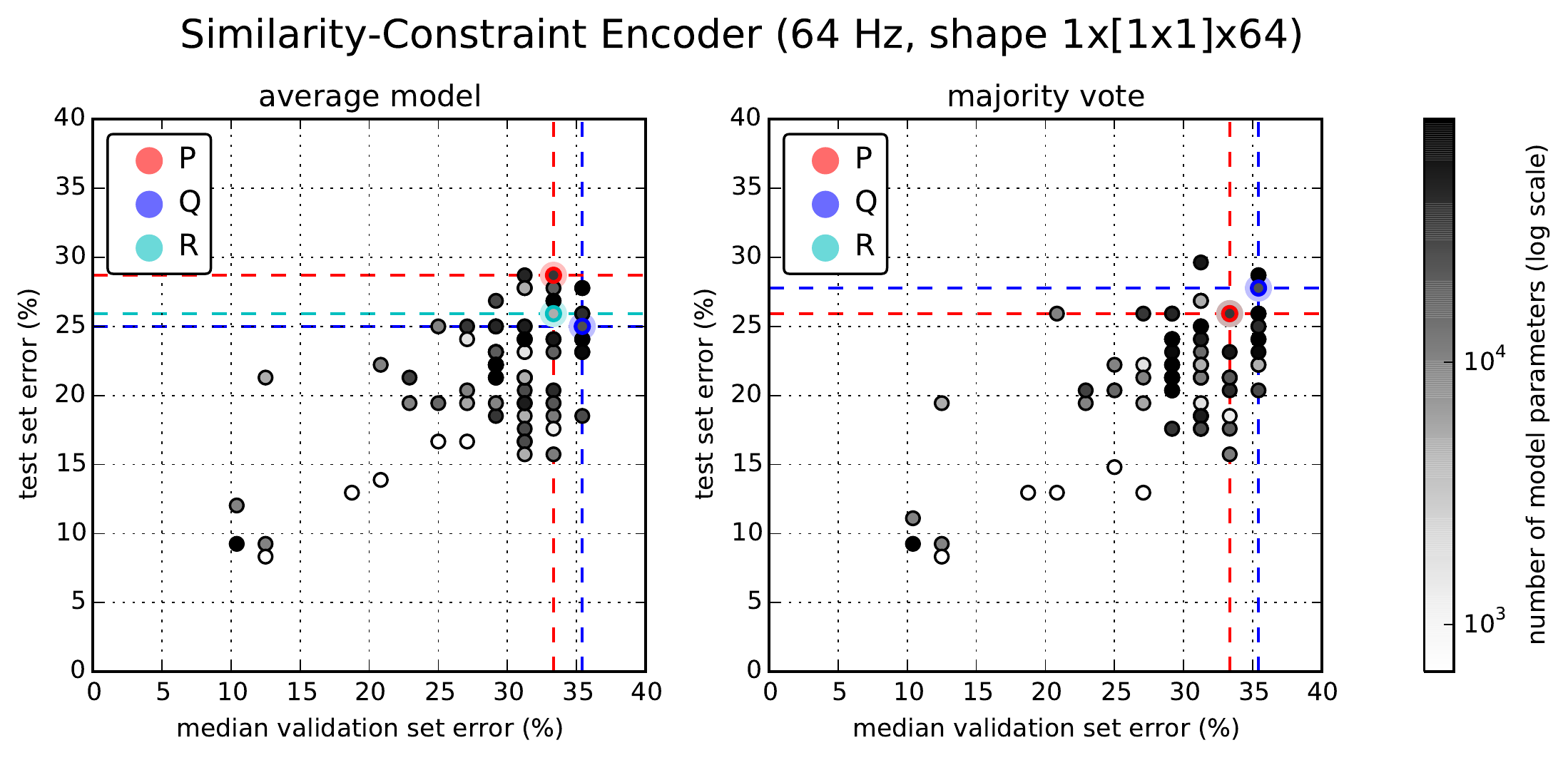}
    \includegraphics[width=\textwidth,keepaspectratio=true]{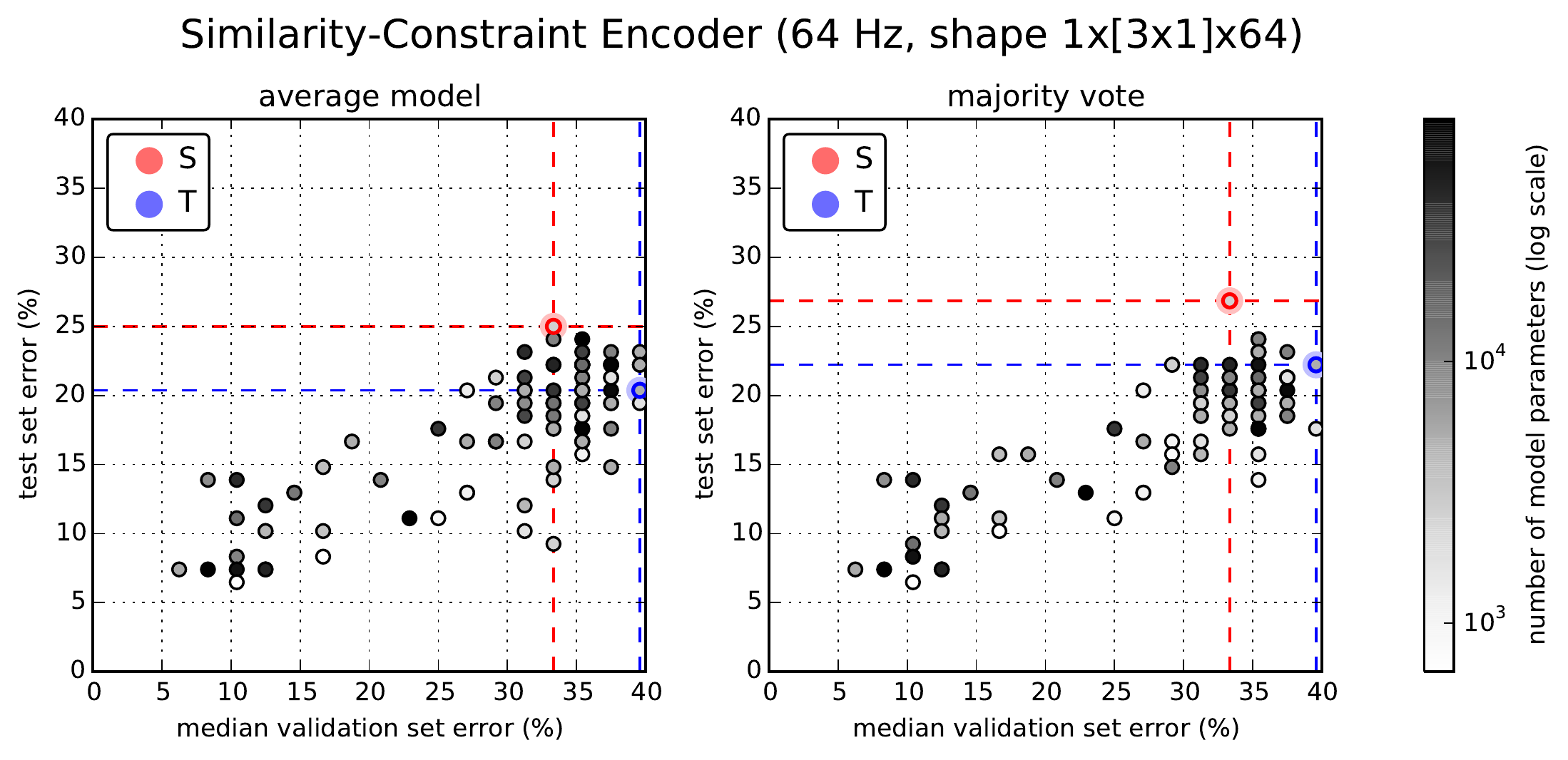}
  \end{center}

  \begin{center}
    \includegraphics[width=\textwidth,keepaspectratio=true]{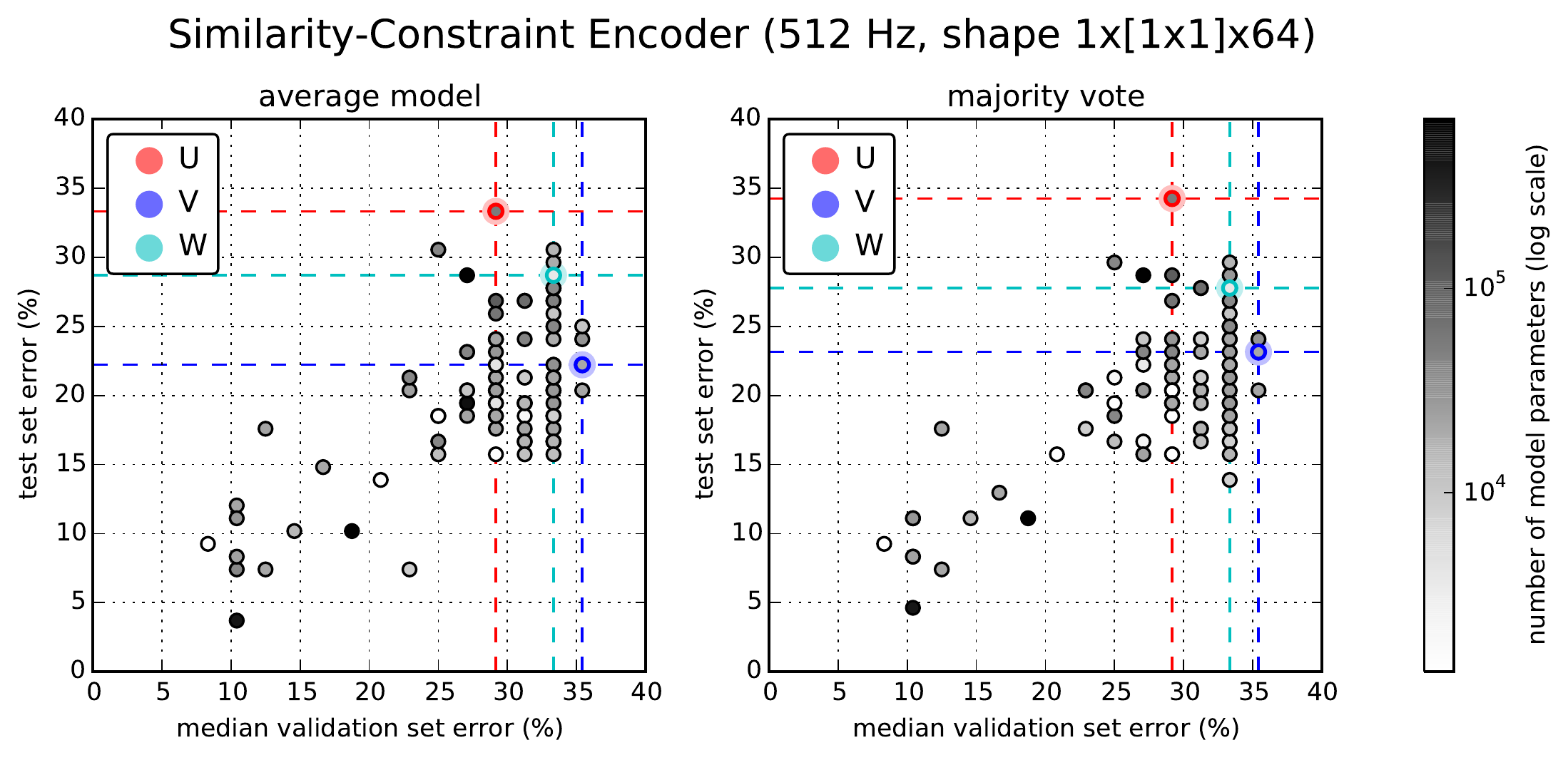}
    \includegraphics[width=\textwidth,keepaspectratio=true]{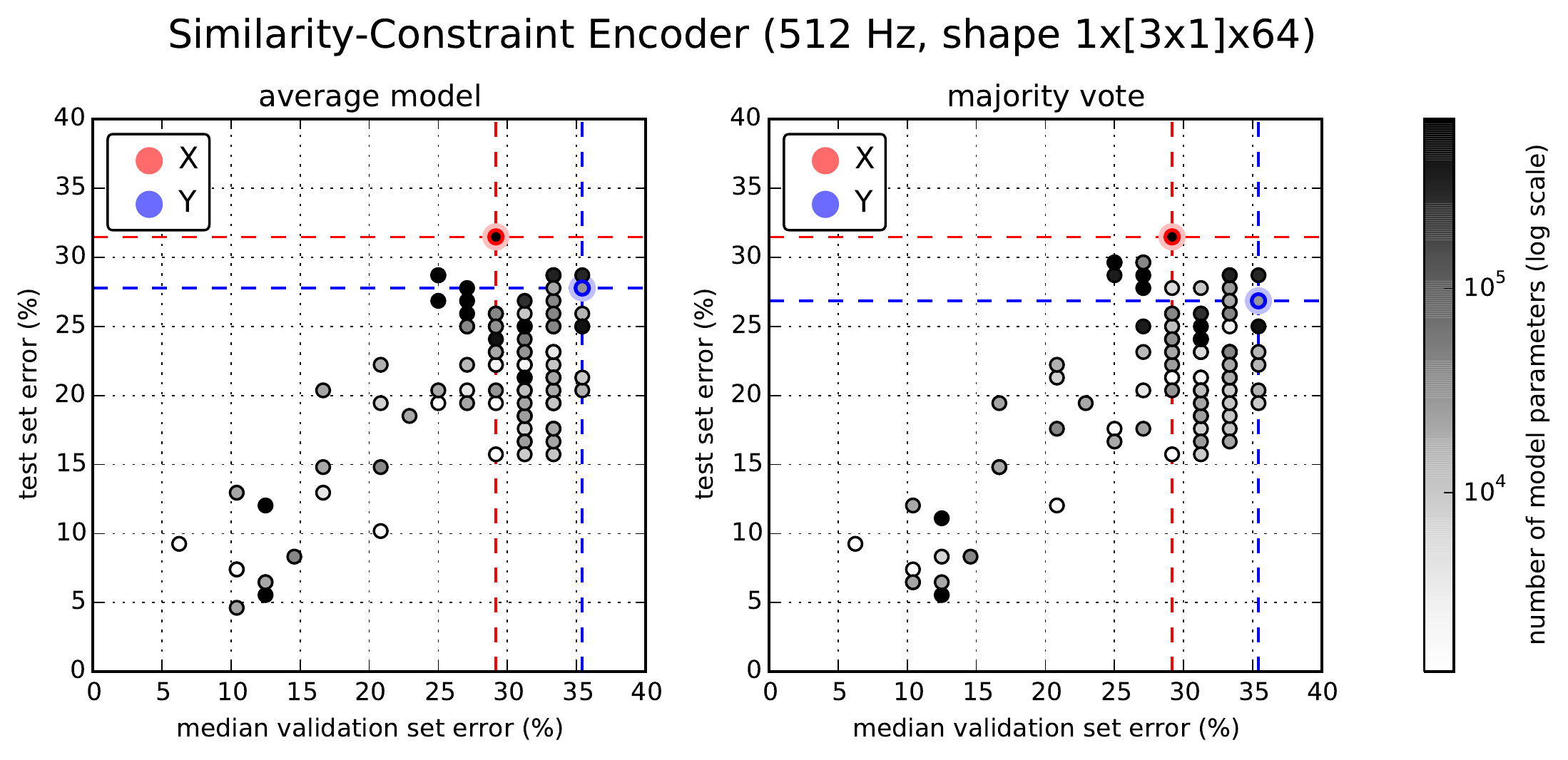}
  \end{center}

\end{document}